%% file: main.tex
\documentclass{article}
\usepackage{globe_preprint,times}
\usepackage[T1]{fontenc}
\input{math_commands.tex}

\usepackage{booktabs}
\usepackage{colortbl}
\definecolor{worse}{RGB}{200,65,47}
\definecolor{better}{RGB}{62,156,139}

\usepackage{multirow}
\usepackage{graphicx}
\usepackage{amsmath,amssymb,mathtools}
\usepackage{microtype}
\usepackage{algorithm}
\usepackage{algorithmic}
\usepackage{float}
\usepackage{mdframed}
\usepackage{url}
\usepackage{hyperref}

\title{GLOBE: Geometry-Led Online Behavioral Early Exit via Unit-Sphere Trajectories}

\author{
Xinyue Peng, Zhenlin Liang, Meng Cai, Hua Wang \\
Moore Threads \\
\texttt{meilong@mthreads.com}
}

\hypersetup{
  pdftitle={GLOBE: Geometry-Led Online Behavioral Early Exit via Unit-Sphere Trajectories},
  pdfauthor={Xinyue Peng, Zhenlin Liang, Meng Cai, Hua Wang},
  bookmarksnumbered=true,
  breaklinks=true
}

\begin{document}
\maketitle

\begin{abstract}
Chain-of-thought (CoT) reasoning can improve problem solving, but models often keep generating after the correct answer is already reachable; the extra steps raise serving costs and may replace a correct intermediate result with an incorrect one. We propose GLOBE (Geometry-Led Online Behavioral Early Exit), an online, training-free algorithm that terminates the model's CoT process early. GLOBE embeds each completed reasoning segment as a point on a unit sphere and reads the shape of the resulting semantic trajectory. GLOBE first applies a trial phase that generates short trial answers in a separate branch, stopping the main stream on an accepted trial after checking confidence against a threshold. If generation continues, a trajectory phase stops the stream once this path repeatedly reverses direction with little net progress. On five benchmarks including mathematical reasoning, tool calling, and agent memory, GLOBE cuts output tokens by 49.7\% and active end-to-end wall time (E2E) by 51.0\% across two models, and achieves a five-task average of 80.58\%, comparable to Full CoT.
\end{abstract}

\input{sections/introduction}
\input{sections/related_work}
\input{sections/method}
\input{experiments}
\input{conclusion}

\clearpage
\subsection*{AI use statement}
In this work, we use generative AI tools to aid and polish writing, including drafting and editing prose, translation, and maintaining the \LaTeX{} source and bibliography. Generative AI tools also assist with implementing and debugging experiment code and providing feedback on experimental design. We have reviewed all AI-assisted work, and we take responsibility for the final content of this work, including text, claims, and artifacts produced with the aid of generative AI.

\subsection*{Reproducibility statement}
The method, controller pseudocode, dataset contracts, and metric definitions are described in Sections~\ref{sec:method}--\ref{sec:experiments} and the appendices.

\bibliographystyle{globe_preprint}
\bibliography{references}

\clearpage
\appendix
\input{algorithm_appendix}
\input{theory_appendix}
\input{experiment_contracts_appendix}
\input{exit_appendix}
\input{mechanism_audit_appendix}
\input{case_studies_appendix}
\end{document}

%% file: math_commands.tex
\usepackage{amsmath,amsfonts,bm}

\def\eqref#1{equation~\ref{#1}}

\def\1{\bm{1}}

\DeclareMathAlphabet{\mathsfit}{\encodingdefault}{\sfdefault}{m}{sl}
\SetMathAlphabet{\mathsfit}{bold}{\encodingdefault}{\sfdefault}{bx}{n}



%% file: sections/introduction.tex
\section{Introduction}
\label{sec:introduction}

Allocating additional test-time computation to chain-of-thought reasoning can
substantially improve complex decision making and problem solving in language
models \citep{wei2022chain,wang2023selfconsistency,snell2024scaling}. However,
longer traces have uncertain marginal value while imposing costs in token
throughput, request concurrency, and end-to-end latency. Long-form reasoning
models often keep generating after the correct answer is already reachable, and
the extra steps can replace a correct intermediate result with an incorrect
one, a behavior known as overthinking
\citep{chen2024overthinking,wu2025more,han2026jet,sui2025stop}. Early exit is
therefore an online allocation decision: from the current prefix, an inference
system must decide whether to stop or purchase more computation without
observing the future continuation.

Early-exit methods monitor different signals. Reflection-marker methods trigger
trial answers and stop when local confidence passes an acceptance test
\citep{yang2025deer}; output-distribution methods compare continuing with
emitting a thinking-close token at syntactic boundaries
\citep{oh2025thinkbrake,wei2026rcpd}; answer-convergence methods wait for
sampled answers to stabilize \citep{liu2025answer}; and other work inspects
step-level redundancy, confidence dynamics, sufficiency, or internal
activations
\citep{min2026puma,hosseini2026codestop,sun2026refrain,xiang2026dtsr,liu2026neat}.
A separate line of work shortens reasoning by training with length-aware
objectives or by fixing a decoding budget
\citep{arora2025training,aggarwal2025l1,muennighoff2025s1}. These signals are
not interchangeable: a trial tests whether the current prefix is ready to answer,
whereas a long trace can keep changing even after local confidence is already high.

\begin{figure}[t]
  \centering
\includegraphics[width=1.00\textwidth]{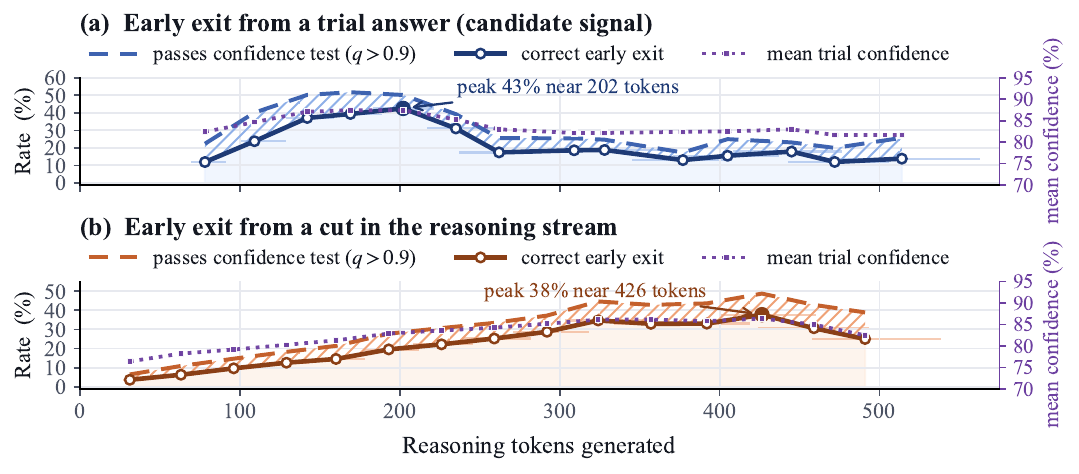}
  \caption{Two mechanisms for triggering an early exit: a trial answer at a candidate
  signal (a) and a cut of the reasoning stream (b).}
  \label{fig:two-triggers}
\end{figure}

We ground the comparison in measurements by replaying the full chain-of-thought
traces of BFCL v4 Live~\citep{patil2025bfcl} offline and reproducing two types of
trigger points: trial answers launched at candidate signals, such as a reflection
marker like \texttt{Wait}; and cuts of the reasoning stream, which is split into
segments with a stop triggered at every cut. For each trigger point we record the
same three quantities: whether the answer reaches high confidence (mean token
probability $q>0.9$), whether that answer is correct, and the average confidence over
those points. Figure~\ref{fig:two-triggers} shows that the two families are offset in
time: trials at candidate signals peak early in the trace and then decay, whereas cuts
of the reasoning stream keep improving until late in the trace. Confidence decides
whether a trial answer is accepted, not when the trial signal stops being useful.
The open question is therefore
whether a controller can tell, during generation, whether the reasoning is still
making progress or is looping back on itself.

To this end, we propose \emph{GLOBE}, a training-free online algorithm over a
single reasoning stream. It launches
short trial answers at candidate signals and embeds each completed segment, so
that generation traces a path on a unit sphere. Early in the trace, a confident
trial answer ends generation; later, the trajectory decides, and the stream
stops once it keeps retracing what it has already covered.

We make three contributions:
\begin{itemize}
    \item We observe that trial-answer stopping and reasoning-stream-cut
    stopping are effective in different regions of the trace: trial answers
    work best early and decay with depth, whereas stream cuts improve later.
    GLOBE therefore introduces a phase switch rather than concatenating them:
    trial confidence asks whether the current prefix is answer-ready, while
    spherical trajectory geometry asks whether reasoning is still advancing.
    \item We introduce a spherical trajectory representation that transforms
    abstract semantic progress into measurable geometric properties. Each
    completed segment is embedded and normalized to a point on the unit sphere,
    where net displacement signals progress and repeated directional reversals
    with little net displacement indicate repetition. The trajectory criterion relies only
    on the angles between segment embeddings. It therefore requires neither
    complex semantic evaluation nor language-specific vocabulary or syntax.
\item \textsc{GLOBE} cuts output tokens and E2E time by about half while keeping its frequent early exits reliable. Across two models and five benchmarks for tasks such as mathematical reasoning, tool calling, and agent memory, \textsc{GLOBE} averages 80.58\% on the five-task score, comparable to 80.04\% for Full CoT, while cutting output tokens by 49.7\% and active E2E time by 51.0\%. On Qwen's GSM8K analysis runs, early exits correct 29.8\% of Full CoT errors while making 0.94\% of Full CoT-correct answers incorrect.
\end{itemize}

%% file: sections/related_work.tex
\section{Related Work}
\label{sec:related_work}

\paragraph{Efficient reasoning and test-time compute.}
Chain-of-thought and test-time scaling improve difficult reasoning by allocating
more inference computation~\citep{wei2022chain,snell2024scaling}, but longer
traces may repeat established conclusions or overturn a correct intermediate
answer~\mbox{\citep{chen2024overthinking,wu2025more}}.  One line of work intervenes at
training time, using supervised or reinforcement learning with length-aware
objectives so that the model itself emits shorter
traces~\citep{arora2025training,aggarwal2025l1,han2026jet}.  Another leaves the
weights untouched and controls length at inference time, through a fixed budget,
budget forcing, concise prompts, or a non-thinking
mode~\citep{muennighoff2025s1,ma2025reasoning}; \citet{sui2025stop} survey this
space.  These approaches control how much reasoning is generated, whereas we
study a training-free online algorithm that decides, from the current
instance's evolving history, whether to purchase more computation.

\paragraph{Online early stopping from local evidence.}
Inference-time methods commonly inspect intermediate answers or decoding
signals.  Answer Convergence stops when predictions from partial traces
stabilize~\citep{liu2025answer}. Certaindex estimates the stability and certainty of intermediate answers to adapt compute~\citep{fu2024certaindex}, and trained hidden-state probes predict answer correctness to support early exiting~\citep{zhang2025reasoning}. DEER uses reflection markers to propose trial
answers and exits when their token confidence is high~\citep{yang2025deer};
ThinkBrake instead monitors the margin between the best continuation token and
the thinking-close token~\citep{oh2025thinkbrake}, while RCPD tracks the rank of
that token at sentence boundaries and force-closes the thinking segment once one
of a few rules distilled from offline reasoning-dynamics analysis is
satisfied~\citep{wei2026rcpd}.  DTSR prompts the model to assess thought
sufficiency at reflection points~\citep{xiang2026dtsr}, while CoDE-Stop and
REFRAIN use confidence or redundancy dynamics over multiple
checkpoints~\citep{hosseini2026codestop,sun2026refrain}.  GLOBE uses such pointwise local evidence only in its
trial phase.

\paragraph{Trajectory-level and deployable controllers.}
Representation-level methods move beyond a single answer score.  PUMA detects
semantic redundancy between reasoning steps and verifies candidate exits with
an answer probe~\citep{min2026puma}; NEAT monitors neuron activation dynamics
without answer rollouts~\citep{liu2026neat}.  Both compare representations
locally, at a step or a pair of steps, rather than describing the shape of the
accumulated trace. Latent-Trajectory signals summarize traces from the
model's hidden states, but select among sampled traces
rather than stop a single one~\citep{vilas2026tracing}.  Deployment adds a second requirement: in tool-using and
agentic settings a stop must preserve the parser-visible action and cross-turn
state~\citep{yao2023react,schick2023toolformer,patil2025bfcl}, and the cost of a
controller includes the trial requests and repeated prefix computation it adds,
not only the tokens it retains.  GLOBE addresses both: it stops a single stream online from the
geometry of its accumulated text-embedding trajectory, and it reports active E2E time alongside
retained tokens under each task's native protocol.

%% file: sections/method.tex
\section{Methodology}
\label{sec:method}

\subsection{GLOBE Overview}
\label{sec:method-overview}

\begin{figure}[!t]
  \centering
  \includegraphics[width=0.95\textwidth]{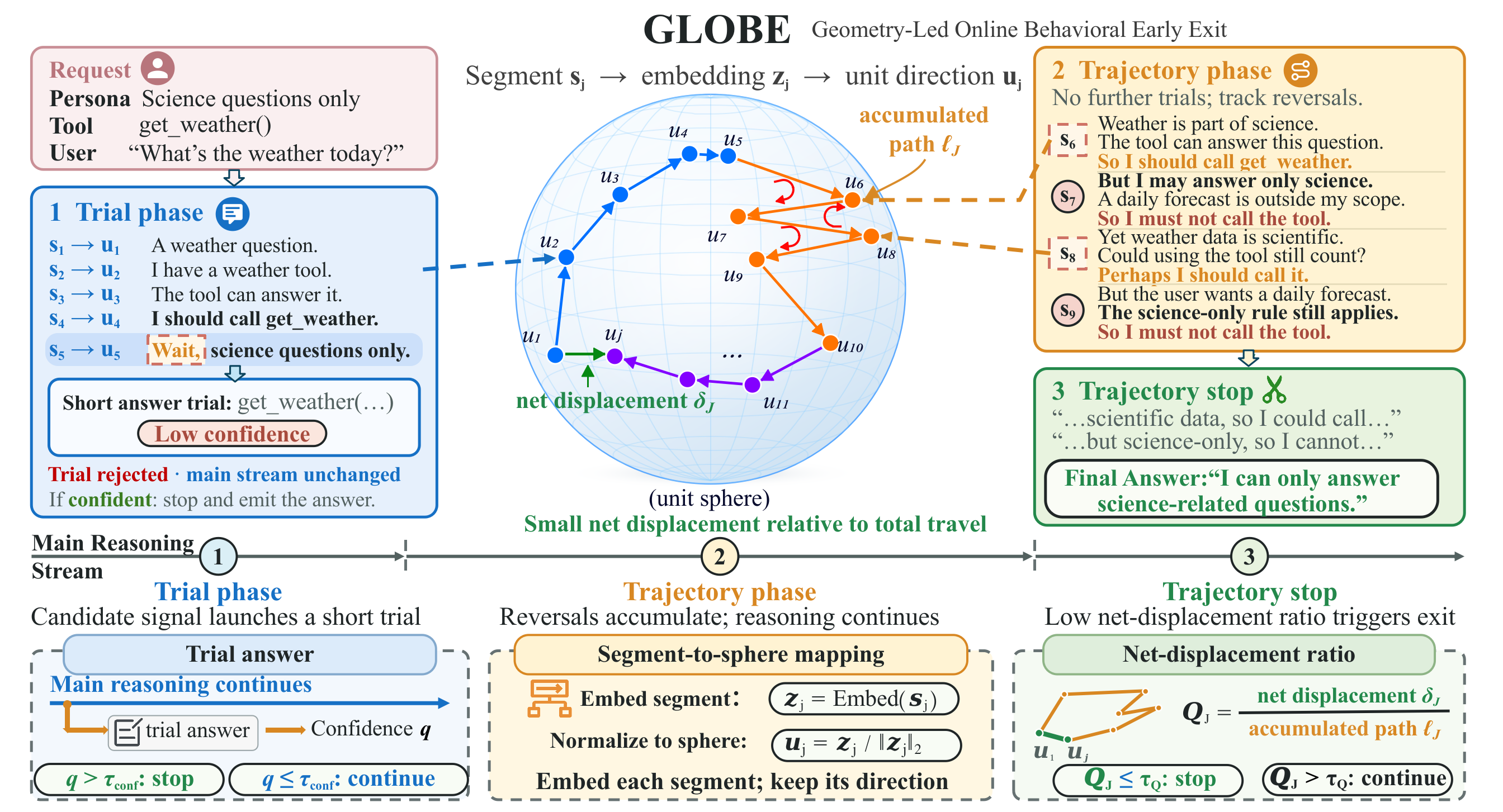}
  \caption{Overview of GLOBE, a training-free online algorithm over a single reasoning stream, in three steps. The example is a tool-calling irrelevance case, as the persona permits only science questions and the weather tool should not be called. In the trial phase, a \texttt{get\_weather} trial answer at \texttt{Wait} is rejected for low confidence. In the trajectory phase, the reasoning wavers over calling the tool. Once reversals, marked by the red arrows, accumulate with little net displacement, further reasoning is redundant, so the trajectory stop ends the stream and the final answer declines the request.}
  \label{fig:globe-overview}
\end{figure}

Figure~\ref{fig:globe-overview} gives an overview of GLOBE, which embeds each completed reasoning segment as a point on a unit sphere and reads the shape of the resulting semantic trajectory. The trial phase generates short trial answers in a separate branch at candidate signals such as \texttt{Wait} and checks their confidence against a threshold. A highly confident trial answer is accepted and stops the main stream; otherwise, it is discarded and the main stream keeps generating. A reversal is a step of the trajectory that moves partly back along the previous step. Once reversals have accumulated, the stream enters the trajectory phase and stops launching trials. In this phase, the trajectory stop ends the stream once the trajectory repeatedly reverses direction with little net progress, that is, a small net displacement relative to its path length. The reasoning is then circling over content it has already covered. After an accepted trial or a trajectory stop, GLOBE generates the final answer from the retained prefix. If neither occurs, the model ends its reasoning naturally and then answers.

\subsection{Candidate-Signal-Triggered Trial Answers}
\label{sec:method-trial}

The trial phase aims to decide whether the current reasoning prefix $r_{\le t}$ already suffices to answer. This prefix consists of the reasoning tokens generated up to step $t$. The decision is made before reasoning ends. For this purpose, GLOBE launches short trial answers at sparse candidate signals. It uses their confidence as the criterion.

The controller supports two types of candidate signals. The first type consists of the two most representative textual markers in the reasoning text. These markers are \texttt{Wait} and \texttt{Actually}. \texttt{Wait} marks a pause to check the current reasoning. \texttt{Actually} marks a re-verification of it. The second type is the task-native thinking-close marker \texttt{</think>}. The controller does not wait for this marker to be generated. Instead, it detects the marker as soon as it enters the top-5 next-token candidates. Thus, the reasoning stream is not closed early. The fact that the model is ready to answer becomes available as a signal, and the reasoning can still continue.

Trial answers are generated in a separate branch, leaving the main reasoning stream uninterrupted. When a signal triggers, the controller freezes the exact prefix, appends a prompt $g$ that elicits a direct answer, and issues one trial request. The request returns a trial answer $\tilde y$ of length $m\le\tau_{\mathrm{trial}}$; the model's predictions are sharply concentrated when the prefix suffices and more dispersed otherwise \citep{yang2025deer}. Let $x$ be the task input, $\mathcal V$ the vocabulary, and $\mathcal H_\theta$ the LM head with parameters $\theta$. The logits at position $i$ ($1\le i\le m$, with the preceding trial tokens $\tilde y_{<i}$) and their largest entry are
\[
\boldsymbol o_i=\mathcal H_\theta\bigl(x,r_{\le t},g,\tilde y_{<i}\bigr),
\qquad
o_i^{\star}=\max_{v\in\mathcal V}o_{i,v},
\]
where $o_i^{\star}$ corresponds to the token chosen by greedy decoding. The confidence of a trial is the mean over positions of the softmax probability of that token:
\begin{equation}
q=\frac{1}{m}\sum_{i=1}^{m}\frac{\exp\bigl(o_i^{\star}\bigr)}{\sum_{v\in\mathcal V}\exp\bigl(o_{i,v}\bigr)}.
\label{eq:confidence}
\end{equation}
A trial is accepted when $q$ exceeds the confidence threshold $\tau_{\mathrm{conf}}$. In this case, the prefix is deemed sufficient to answer. The decision rests on the model's own predictive distribution, not on the correctness of the answer.

After a trial is accepted, the system runs the continuation procedure. First, it cancels the main reasoning stream and discards everything after the frozen point. Second, it re-issues the retained prefix to the same serving endpoint. Thus, its prefix cache is reused. Third, it appends \texttt{</think>} and generates the final answer under the original prompt and native parsing rules. A rejected trial is discarded and leaves the main reasoning stream unaffected. Trial requests are accounted for separately. Appendices~\ref{app:implementation-details} and~\ref{app:formula-boundaries} give the exact prefix boundaries, edge cases, and cost accounting.

\subsection{Unit-Sphere Trajectory Representation}
\label{sec:method-sphere}

Fewer trial answers are both confident and correct late in reasoning (Figure~\ref{fig:two-triggers}), so the next signal must come from the reasoning content itself: is the chain advancing, or is it returning to what it has covered? Prior work shows that embeddings encode semantics: the squared norm of a word embedding carries its information gain, a result extending to contextualized embeddings \mbox{\citep{oyama-etal-2023-norm}}, while semantic relations are reflected in the directions of their difference vectors \citep{mikolov-etal-2013-linguistic,ethayarajh-etal-2019-towards}. Since our decision concerns the relation between successive reasoning segments rather than the information gain carried by any one of them, and our encoder's cosine-similarity objective makes that relation angular \citep{zhang2025qwen3embedding}, we normalize each segment embedding onto the unit sphere, so that advance and return are read from the angles alone.

The geometric observer runs alongside the main reasoning stream. The segmenter splits the reasoning text $r$ into segments $s_1,s_2,\ldots$ at sentence-ending punctuation and newlines. It merges adjacent fragments until each completed segment has at least the minimum length $L_{\min}$ (in characters). This prevents over-fragmentation. Each completed segment is then encoded into a semantic vector by the embedding model. Appendix~\ref{app:segmentation} describes the segmenter. Embedding overlaps with generation; its request time is at most $13.6\%$ of per-item wall time in the profiled Qwen runs (Appendix~\ref{app:overhead}). Appendix~\ref{app:encoder} gives encoder details and results with smaller encoders.

Let $J$ be the number of completed segments and $z_j\in\mathbb R^d$ the semantic vector of segment $s_j$, where $d$ is the embedding dimension. We normalize this vector as $u_j=z_j/\|z_j\|_2$, where $\|\cdot\|_2$ denotes the Euclidean norm. The reasoning trajectory is the point sequence $(u_j)_{j=1}^{J}$ on the unit sphere $\mathbb S^{d-1}=\{u\in\mathbb R^d:\|u\|_2=1\}$. Its chordal displacements are $\xi_j=u_{j+1}-u_j$. The controller incrementally maintains the needed entries of the Gram matrix $S=[S_{a,b}]$, with entries $S_{a,b}=\langle u_a,u_b\rangle$ for $a,b\in\{1,\dots,J\}$, where $\langle\cdot,\cdot\rangle$ denotes the Euclidean inner product. The chord length $\|u_a-u_b\|_2=\sqrt{2-2S_{a,b}}$ depends only on the angle. Thus, the statistics below are functions of the Gram matrix $S$ alone. They are invariant under orthogonal transformations of the embedding space and under positive rescaling of the raw embeddings $z_j$. Each new segment therefore requires only $O(d)$ time to update these quantities.

To detect changes of direction, we compute the inner products of consecutive chordal displacements:
\begin{equation}
\kappa_j=\langle \xi_j,\xi_{j+1}\rangle=\tfrac12\bigl(\|u_{j+2}-u_j\|_2^2-\|\xi_j\|_2^2-\|\xi_{j+1}\|_2^2\bigr)=S_{j,j+1}+S_{j+1,j+2}-S_{j,j+2}-1.
\label{eq:gram-reversal}
\end{equation}
Here $\kappa_j$ is defined for $1\le j\le J-2$, and $\kappa_j<0$ means one directional reversal. By the polarization identity, this is equivalent to the reasoning not continuing along its previous direction (Appendix~\ref{app:sphere-theory}). We count reversals by $K_J=\sum_{j=1}^{J-2}\mathbf 1[\kappa_j<0]$. Here, $\mathbf 1[\cdot]$ is the indicator function.

Once the reversals have accumulated, we also need to decide whether the reasoning is going in circles instead of making progress. For this purpose, we first look at how far the trace has traveled and how far it has actually advanced. These distances are the path length and the start-to-end displacement:
\begin{equation}
\ell_J=\sum_{j=1}^{J-1}\|\xi_j\|_2=\sum_{j=1}^{J-1}\sqrt{2-2S_{j,j+1}},\qquad
\delta_J=\|u_J-u_1\|_2=\sqrt{2-2S_{1,J}}.
\label{eq:path-length}
\end{equation}
We define the net-displacement ratio as $Q_J=\delta_J/\ell_J$ when
$\ell_J>0$, and set $Q_J=1$ when $\ell_J=0$. A smaller value means that the
trace returns more relative to the path it has traveled.

\subsection{Phase Switch and Trajectory Stop}
\label{sec:method-phase}

Let $J(t)$ be the number of segments completed by step $t$, and write $K_t=K_{J(t)}$ and $Q_t=Q_{J(t)}$ for the reversal count and the net-displacement ratio at that point. The constants $K_{\mathrm{phase}}$, $K_{\mathrm{osc}}$, and $\tau_Q$ are the phase-switch, oscillation, and net-displacement-ratio thresholds. The early-exit indicator at step $t$ is
\begin{equation}
e_t=\begin{cases}
\mathbf 1[q>\tau_{\mathrm{conf}}], & K_t<K_{\mathrm{phase}}\ \text{and a signal triggers at } t,\\
\mathbf 1[K_t\ge K_{\mathrm{osc}}\ \text{and}\ Q_t\le \tau_Q], & K_t\ge K_{\mathrm{phase}},\\
0, & \text{otherwise}.
\end{cases}
\label{eq:exit-rule}
\end{equation}
GLOBE exits at the first step with $e_t=1$. The first case is the trial phase. Here, the signal is a candidate signal of Section~\ref{sec:method-trial}. Once $K_t$ reaches the phase threshold $K_{\mathrm{phase}}$, the stream enters the trajectory phase of the second case. In this phase, no further trials are launched. Also, no exit is possible until $K_t$ reaches the oscillation threshold $K_{\mathrm{osc}}$. A trajectory stop is placed at the boundary of the latest completed segment. If $e_t=0$ throughout, reasoning ends at the model's own termination. Both early exits execute the continuation procedure of Section~\ref{sec:method-trial}. Appendix~\ref{app:reference-config} lists the operating thresholds.

%% file: experiments.tex
\section{Experiments}
\label{sec:experiments}

\subsection{Experimental setup}
\label{sec:experimental-setup}

\paragraph{Benchmarks and models.} We evaluate five benchmarks: commonsense question answering on CSQA~\mbox{\citep{talmor2019commonsenseqa}}, role-play character fidelity on Character-LLM~\mbox{\citep{shao2023character}}, grade-school mathematics on GSM8K~\mbox{\citep{cobbe2021gsm8k}}, tool calling on BFCL v4 Live~\mbox{\citep{patil2025bfcl}}, and long-horizon agent memory on BFCL v4 Agent Memory~\mbox{\citep{patil2025bfcl}}, whose KV, Vector, and Recursive Sum backends are scored separately. We use two instruction-tuned thinking models from different families and parameter scales, Qwen3.6-35B-A3B~\mbox{\citep{qwen2026qwen36}}, a mixture-of-experts model with 35B total parameters, and Gemma-4-12B-it~\mbox{\citep{gemma2026gemma4}}, which we call Qwen and Gemma. Accuracy or exact match is the native metric of the four accuracy-style benchmarks; for Character-LLM, a fixed judge applies a four-aspect adaptation of the original rubric and we report the resulting normalized acting score (NAS), defined in Appendix~\ref{app:contracts}. We also report average retained output tokens and active task-level end-to-end wall time (E2E).


\begin{table}[!t]
\centering
\caption{Main results. Acc.\ (\%), Tok.\ (average output tokens), and E2E (s)
per task. Agent Memory Acc.\ averages KV, Vector, and Recursive Sum, and its
E2E is their sum. Bold: best Acc., and lowest Tok/E2E among thinking methods.}
\label{tab:all-results}
\small
\setlength{\tabcolsep}{1.0pt}
\renewcommand{\arraystretch}{1.20}
\resizebox{\textwidth}{!}{%
\begin{tabular}{@{}l|ccc|ccc|ccc|ccc|ccc|ccc@{}}
\toprule
\multirow{2}{*}{\textbf{Method}} & \multicolumn{3}{c|}{\textbf{CSQA}} & \multicolumn{3}{c|}{\textbf{Character-LLM}} &
\multicolumn{3}{c|}{\textbf{GSM8K}} & \multicolumn{3}{c|}{\textbf{BFCL v4 Live}} & \multicolumn{3}{c|}{\textbf{Agent Memory}} & \multicolumn{3}{c}{\textbf{Avg.}}\\
& \textbf{Acc$\uparrow$} & \textbf{Tok$\downarrow$} & \textbf{E2E$\downarrow$} & \textbf{Acc$\uparrow$} & \textbf{Tok$\downarrow$} & \textbf{E2E$\downarrow$} & \textbf{Acc$\uparrow$} & \textbf{Tok$\downarrow$} & \textbf{E2E$\downarrow$} & \textbf{Acc$\uparrow$} & \textbf{Tok$\downarrow$} & \textbf{E2E$\downarrow$} & \textbf{Acc$\uparrow$} & \textbf{Tok$\downarrow$} & \textbf{E2E$\downarrow$} & \textbf{Acc$\uparrow$} & \textbf{Tok$\downarrow$} & \textbf{E2E$\downarrow$}\\
\midrule
\midrule
\multicolumn{18}{l}{\textbf{Qwen3.6-35B-A3B}}\\
Nothink & 83.8 & 601 & 1062 & 90.8 & 124 & 1323 & 96.3 & 189 & 575 & 77.9 & 103 & 1015 & 34.5 & 763 & 1556 & 76.66 & 356 & 1106\\
\rowcolor{yellow!10}
\textbf{Full CoT} & 84.7 & 2065 & 5376 & 94.4 & 1368 & 7775 & \textbf{96.7} & 1761 & 3475 & \textbf{81.3} & 962 & 4888 & 55.0 & 9549 & 16304 & 82.42 & 3141 & 7564\\
DEER & 81.4 & 1584 & 3090 & \textbf{95.2} & 1233 & 9515 & 96.1 & 1240 & 2631 & 77.3 & 581 & 4964 & 55.4 & 6839 & 14434 & 81.08 & 2295 & 6927\\
ThinkBrake & 82.6 & 2044 & 5719 & 94.4 & \textbf{355} & 6458 & 96.1 & 1779 & 5539 & 79.6 & 789 & 4818 & 52.9 & 7417 & 13673 & 81.12 & 2477 & 7241\\
RCPD & 85.1 & \textbf{514} & \textbf{1174} & 93.2 & 399 & \textbf{3424} & 96.4 & \textbf{570} & \textbf{1500} & 65.6 & \textbf{303} & \textbf{2547} & 51.9 & 2171 & \textbf{4656} & 78.44 & \textbf{791} & \textbf{2660}\\
\rowcolor{green!10}
\textbf{GLOBE (ours)} & \textbf{86.4} & 525 & 1290 & 94.5 & 933 & 5389 & 96.5 & 784 & 1969 & 80.0 & 371 & 3071 & \textbf{58.0} & \textbf{2040} & 5730 & \textbf{83.08} & 931 & 3490\\
\midrule
\multicolumn{18}{l}{\textbf{Gemma-4-12B-it}}\\
Nothink & 81.2 & 342 & 837 & 91.7 & 117 & 497 & 94.0 & 160 & 635 & 77.1 & 51 & 888 & 31.8 & 766 & 2683 & 75.16 & 287 & 1108\\
\rowcolor{yellow!10}
\textbf{Full CoT} & \textbf{83.0} & 1939 & 5009 & 91.5 & 699 & 9893 & 94.3 & 1382 & 4047 & 77.6 & 601 & 5782 & 41.9 & 2308 & 7523 & 77.66 & 1386 & 6451\\
DEER & 77.9 & 743 & 2053 & \textbf{94.3} & 646 & 2611 & 88.6 & 1414 & 4846 & 76.4 & 526 & 5600 & 38.7 & 1951 & 7286 & 75.18 & 1056 & 4479\\
ThinkBrake & 82.1 & 1831 & 4792 & 92.8 & \textbf{500} & 7901 & \textbf{95.4} & 1424 & 4322 & 76.7 & 386 & 3693 & 38.3 & 2217 & 6671 & 77.06 & 1272 & 5476\\
RCPD & 82.2 & 1268 & 3624 & 93.9 & 1175 & 5458 & 89.7 & 1066 & 3865 & 77.6 & 466 & 4962 & \textbf{42.6} & 2210 & 7120 & 77.20 & 1237 & 5006\\
\rowcolor{green!10}
\textbf{GLOBE (ours)} & 82.0 & \textbf{689} & \textbf{2034} & 92.4 & 645 & \textbf{2478} & 95.1 & \textbf{717} & \textbf{2306} & \textbf{81.1} & \textbf{306} & \textbf{2744} & 39.8 & \textbf{1708} & \textbf{5442} & \textbf{78.08} & \textbf{813} & \textbf{3001}\\
\bottomrule\end{tabular}%
}
\end{table}

\paragraph{Baselines.}
Nothink disables the thinking channel and answers directly, bounding the cost of
reasoning from below~\mbox{\citep{ma2025reasoning}}. Full CoT leaves thinking
unconstrained and serves as the quality reference.
DEER~\mbox{\citep{yang2025deer}} watches reflection markers, issues a trial
answer at each marker, and exits the stream once that trial answer is confident.
ThinkBrake~\mbox{\citep{oh2025thinkbrake}} uses no trial answers: at sentence
boundaries it compares the log probabilities of continuing and of emitting
\texttt{</think>}, and stops on a margin criterion.
RCPD~\mbox{\citep{wei2026rcpd}} also uses no trial answers and tracks the rank
of \texttt{</think>} at sentence boundaries, closing the thinking segment once
one of a few rules distilled from an offline reasoning-dynamics analysis is
satisfied. Each baseline follows its method-native contract.

\subsection{Main results}
\label{sec:main-results}

\paragraph{Overall performance.}
Table~\ref{tab:all-results} compares GLOBE with the Nothink and Full CoT
controls and with the early-exit baselines of
Section~\ref{sec:experimental-setup}. GLOBE attains the highest five-task
average on both models, reaching 83.08\% on Qwen and 78.08\% on Gemma against
82.42\% and 77.66\% for Full CoT. The reduction holds on every task: averaged over
the ten model--task pairs, output tokens fall by 49.7\% and E2E time by
51.0\%, and the five-task average output-token count falls from 3141 to 931 on
Qwen and from 1386 to 813 on Gemma.

\paragraph{Comparison with early-exit baselines.} DEER loses 5.7 accuracy points on GSM8K and 5.1 on CSQA in Gemma; its stop needs a reflection marker where the trial answer is already decidable, which is not guaranteed across models. ThinkBrake is the most conservative: no task's accuracy deviates from Full CoT by more than 3.6 points, and it stays below Full CoT in both five-task averages. RCPD is the cheapest controller on Qwen in tokens and E2E, averaging 791 tokens and 2660~s, but it sacrifices accuracy: a 78.44\% five-task average, 4.64 points below GLOBE and 15.7 points below Full CoT on BFCL v4 Live, suggesting its rank rule can trigger before the answer settles. No baseline reaches the Full CoT five-task average on either model, while GLOBE exceeds it on both.

\paragraph{GLOBE is stable across models and datasets.} The Nothink--Full CoT gap shows that thinking matters most on Agent Memory, with gaps of 20.5\% and 10.1\% on the two models, while the gap is 0.4\% and 0.3\% on GSM8K and at most 3.6\% on the other three tasks. GLOBE's worst task is 2.1 points below Full CoT, and it spends fewer tokens and less E2E time than Full CoT in all ten model--task cells. The baselines diverge: on the five-task average, DEER falls $1.34$ points below Full CoT on Qwen and $2.48$ below it on Gemma, whereas ThinkBrake falls $1.30$ and $0.60$ points below, respectively. RCPD is the weakest baseline on Qwen yet the strongest on Gemma. RCPD's savings are equally model-dependent, removing 71.9\% of output tokens on Qwen but only 3.2\% on Gemma. \subsection{Ablation study} \label{sec:component-ablation}

Table~\ref{tab:remove-one} removes one component of GLOBE at a time and
compares each variant with the GLOBE row of Table~\ref{tab:all-results}.
Removing any component lowers the three-task average accuracy on both models,
by up to $11.3$ points. The three tasks span the difficulty tiers of
Section~\ref{sec:main-results}: GSM8K depends on thinking least, Agent
Memory most, and BFCL v4 Live in between; Appendix
Tables~\ref{tab:csqa-ablation} and~\ref{tab:character-ablation} extend the sweep
to CSQA and Character-LLM.

\begin{table}[!t]
\centering
\caption{Component ablation: accuracy (\%) and average output tokens, reporting changes against GLOBE (Table~\ref{tab:all-results}). Parentheses give accuracy differences in points and relative token changes (\%), computed before rounding. The Average token column gives the geometric mean of the token ratios. \textcolor{worse}{red} is worse than GLOBE, \textcolor{better}{green} better, and gray a tie.}
\label{tab:remove-one}
\small
\setlength{\tabcolsep}{2.5pt}
\renewcommand{\arraystretch}{1.20}
\resizebox{\textwidth}{!}{%
\begin{tabular}{@{}l|cc|cc|cc|cc@{}}
\toprule
\multirow{2}{*}{\textbf{Method}} & \multicolumn{2}{c|}{\textbf{GSM8K}} & \multicolumn{2}{c|}{\textbf{BFCL v4 Live}} & \multicolumn{2}{c|}{\textbf{Agent Memory}} & \multicolumn{2}{c}{\textbf{Average}}\\
& \textbf{Accuracy$\uparrow$} & \textbf{Avg Token$\downarrow$} & \textbf{Accuracy$\uparrow$} & \textbf{Avg Token$\downarrow$} & \textbf{Accuracy$\uparrow$} & \textbf{Avg Token$\downarrow$} & \textbf{Accuracy$\uparrow$} & \textbf{Avg Token$\downarrow$}\\
\midrule
\multicolumn{9}{l}{\textbf{Qwen3.6-35B-A3B}}\\
\rowcolor{green!10}
\textbf{GLOBE (ours)} & 96.5 & 784 & 80.0 & 371 & 58.0 & 2040 & 78.2 & 1.00$\times$\\
\quad w/o trajectory stop & 96.1 {\scriptsize\color{worse}($-0.5$)} & 1529 {\scriptsize\color{worse}($+95$\%)} & 80.0 {\scriptsize\color{gray}($+0.0$)} & 757 {\scriptsize\color{worse}($+104$\%)} & 46.9 {\scriptsize\color{worse}($-11.1$)} & 9181 {\scriptsize\color{worse}($+350$\%)} & 74.3 {\scriptsize\color{worse}($-3.8$)} & {\color{worse}2.62$\times$}\\
\quad w/o phase switch & 96.3 {\scriptsize\color{worse}($-0.2$)} & 676 {\scriptsize\color{better}($-14$\%)} & 79.5 {\scriptsize\color{worse}($-0.4$)} & 336 {\scriptsize\color{better}($-9$\%)} & 50.1 {\scriptsize\color{worse}($-7.9$)} & 2551 {\scriptsize\color{worse}($+25$\%)} & 75.3 {\scriptsize\color{worse}($-2.9$)} & {\color{better}0.99$\times$}\\
\quad w/o trial answers & 95.8 {\scriptsize\color{worse}($-0.8$)} & 959 {\scriptsize\color{worse}($+22$\%)} & 79.7 {\scriptsize\color{worse}($-0.3$)} & 412 {\scriptsize\color{worse}($+11$\%)} & 56.1 {\scriptsize\color{worse}($-1.9$)} & 2518 {\scriptsize\color{worse}($+23$\%)} & 77.2 {\scriptsize\color{worse}($-1.0$)} & {\color{worse}1.19$\times$}\\
\quad w/o $\tau_Q$ gate & 96.2 {\scriptsize\color{worse}($-0.3$)} & 623 {\scriptsize\color{better}($-21$\%)} & 79.1 {\scriptsize\color{worse}($-0.9$)} & 349 {\scriptsize\color{better}($-6$\%)} & 55.7 {\scriptsize\color{worse}($-2.3$)} & 2224 {\scriptsize\color{worse}($+9$\%)} & 77.0 {\scriptsize\color{worse}($-1.2$)} & {\color{better}0.93$\times$}\\
\midrule
\multicolumn{9}{l}{\textbf{Gemma-4-12B-it}}\\
\rowcolor{green!10}
\textbf{GLOBE (ours)} & 95.1 & 717 & 81.1 & 306 & 39.8 & 1708 & 72.0 & 1.00$\times$\\
\quad w/o trajectory stop & 62.2 {\scriptsize\color{worse}($-32.9$)} & 3877 {\scriptsize\color{worse}($+441$\%)} & 75.5 {\scriptsize\color{worse}($-5.6$)} & 574 {\scriptsize\color{worse}($+88$\%)} & 44.5 {\scriptsize\color{better}($+4.7$)} & 2044 {\scriptsize\color{worse}($+20$\%)} & 60.7 {\scriptsize\color{worse}($-11.3$)} & {\color{worse}2.30$\times$}\\
\quad w/o phase switch & 90.1 {\scriptsize\color{worse}($-5.0$)} & 1065 {\scriptsize\color{worse}($+49$\%)} & 75.7 {\scriptsize\color{worse}($-5.4$)} & 294 {\scriptsize\color{better}($-4$\%)} & 36.6 {\scriptsize\color{worse}($-3.2$)} & 1587 {\scriptsize\color{better}($-7$\%)} & 67.5 {\scriptsize\color{worse}($-4.5$)} & {\color{worse}1.10$\times$}\\
\quad w/o trial answers & 93.1 {\scriptsize\color{worse}($-2.0$)} & 832 {\scriptsize\color{worse}($+16$\%)} & 77.0 {\scriptsize\color{worse}($-4.1$)} & 283 {\scriptsize\color{better}($-8$\%)} & 36.1 {\scriptsize\color{worse}($-3.7$)} & 1551 {\scriptsize\color{better}($-9$\%)} & 68.8 {\scriptsize\color{worse}($-3.2$)} & {\color{better}0.99$\times$}\\
\quad w/o $\tau_Q$ gate & 90.4 {\scriptsize\color{worse}($-4.6$)} & 973 {\scriptsize\color{worse}($+36$\%)} & 76.0 {\scriptsize\color{worse}($-5.1$)} & 289 {\scriptsize\color{better}($-6$\%)} & 36.3 {\scriptsize\color{worse}($-3.4$)} & 1523 {\scriptsize\color{better}($-11$\%)} & 67.6 {\scriptsize\color{worse}($-4.4$)} & {\color{worse}1.05$\times$}\\
\bottomrule
\end{tabular}%
}
\end{table}

\paragraph{The trajectory stop controls token cost.}
Removing it is the most costly ablation: all six token cells rise, by a factor
of $2.62$ on Qwen and $2.30$ on Gemma. Without it, the Agent Memory output on Qwen grows from 2040 to 9181 tokens. The largest observed accuracy gap is $32.9$ points on GSM8K in Gemma, followed by $11.1$ on Agent Memory in Qwen; Gemma Agent Memory gains $4.7$ points, so the trajectory stop contributes more than any other component. Appendix~\ref{app:gemma-gsm-protocol} shows that the Gemma GSM8K loss comes from traces that never close within the token budget. The criterion
matters wherever the trace stalls: the stream closes once the trajectory
reverses repeatedly with little net displacement. Geometry judges whether the
trajectory still advances without certifying correctness, so dropping the stop
lets generation continue into a post-retrieval restatement that resembles
oscillation.

\paragraph{Removing the phase switch costs accuracy.} The phase switch is one-way: once the cumulative reversal count reaches $K_{\mathrm{phase}}$, the controller permanently retires trial answers and only the trajectory stop can close the stream; removing it lets trial answers return in the trajectory phase. It costs an average of $2.9$ accuracy points on Qwen and $4.5$ on Gemma. On Qwen the loss concentrates on Agent Memory, which drops by $7.9$ points, while GSM8K and BFCL v4 Live save at most $14\%$ of tokens for at most $0.4$ accuracy points; on Gemma every task loses accuracy, by $3.2$ to $5.4$ points. Removing it changes average token use by at most $10\%$ yet consistently lowers the three-task average, so the savings cannot offset the accuracy loss.

\paragraph{Trial answers and the $\tau_Q$ gate are jointly required.} Trial answers freeze the prefix at a candidate signal such as \texttt{Wait} and generate a short answer in a separate branch, accepted only when its confidence exceeds a threshold; the $\tau_Q$ gate vetoes trajectory-stop candidates when $Q_t>\tau_Q$. Removing either component lowers accuracy in all twelve model--task combinations, by up to $5.1$ points, and the three-task averages by up to $4.4$. The cheapest Qwen variant drops the $\tau_Q$ gate and uses $0.93\times$ the tokens of GLOBE, but loses accuracy on all three tasks; on Gemma, dropping either component saves up to $11\%$ of tokens on BFCL v4 Live and Agent Memory for up to $5.1$ accuracy points. No cell is simultaneously cheaper and more accurate than GLOBE: the trials test whether the current prefix already supports an answer, and the gate vetoes stops that would come too early.

\subsection{Stop reliability}
\label{sec:stop-reliability}

\begin{figure}[t]
  \centering
\includegraphics[width=1.00\textwidth]{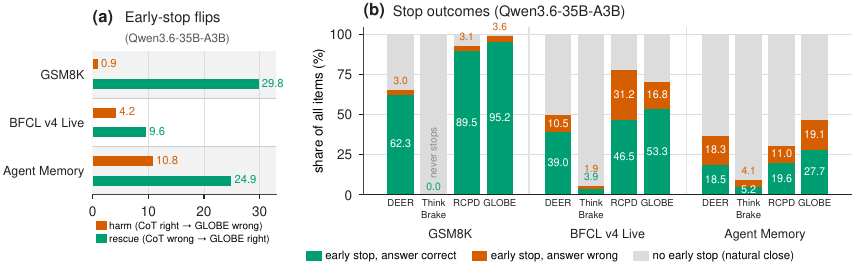}
  \caption{Stop reliability on separate analysis runs. (a)~Harm and rescue of GLOBE early stops versus
  Full CoT on GSM8K, BFCL v4 Live, and Agent Memory. (b)~Stop outcomes
  per controller: stopped and correct (green), stopped and wrong (orange), and
  natural close (grey).}
  \label{fig:stop}
\end{figure}

\paragraph{Early stopping can recover some Full CoT errors.}
Figure~\ref{fig:stop}a joins Full CoT with GLOBE by item: harm is the share of
Full CoT-correct items that GLOBE stops early and answers incorrectly, and
rescue is the share of Full CoT-wrong items that GLOBE stops early and answers correctly. GSM8K is
the least thinking-dependent task, so answers are usually already determined:
harm is rare at $0.94\%$ and rescue reaches $29.8\%$. On BFCL v4 Live, a
tool call must keep its choice, call structure and arguments consistent, so a
settled-looking prefix can still become a different call; harm is $4.17\%$ and
rescue $9.56\%$. On Agent Memory the two rates are at $10.8\%$ and
$24.9\%$, because the stateful task also carries the run-to-run divergence of
the memory contents into the join.

\paragraph{GLOBE converts the most early stops into correct answers.}
Figure~\ref{fig:stop}b splits each Qwen task into correct early stops in green,
incorrect early stops in orange and items left to close naturally in grey, so
the green band is the share of items stopped early and answered correctly. GLOBE
has the largest green band on all three tasks, at $95.2\%$ on GSM8K, $53.3\%$ on
BFCL v4 Live and $27.7\%$ on Agent Memory. DEER requires a reflection marker
plus a confident trial, capping its green band at $62.3\%$ on
GSM8K and $39.0\%$ on BFCL v4 Live. ThinkBrake triggers
only when closing the segment comes within a fixed margin of
continuing; on GSM8K the closing decision stays far behind continuing at every
sentence boundary, so it is marked \textit{never stops}. On BFCL v4 Live it reaches only $3.9\%$. RCPD
stops more than DEER or ThinkBrake but converts less: on GSM8K and BFCL v4 Live
its green bands are $89.5\%$ and $46.5\%$, against $95.2\%$ and $53.3\%$ for
GLOBE, and only $59.8\%$ of its BFCL stops are correct. Under GLOBE, $16.8\%$ of BFCL v4 Live items are stopped early with a wrong answer, against $31.2\%$ under RCPD. On Agent Memory it
stops only $30.5\%$ of items for a $19.6\%$ green band against $27.7\%$. RCPD
thus shows both failure modes of a surface trigger: low conversion on BFCL v4
Live and rare stops on Agent Memory.


\subsection{Operating-point evidence}
\label{sec:evidence-horizon-selection}

\begin{figure}[!t]
  \centering
  \includegraphics[width=1.00\textwidth]{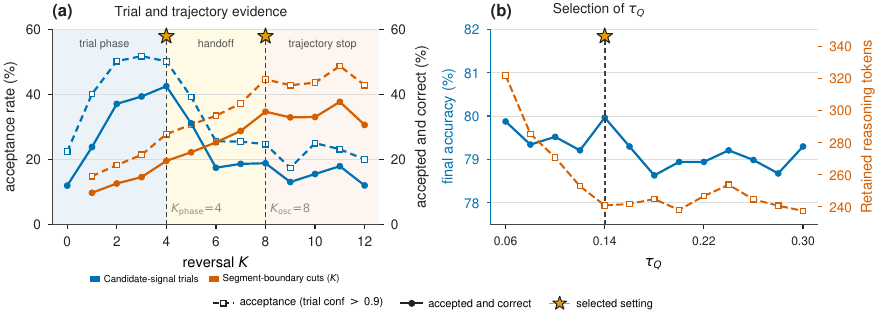}
  \caption{Operating-point evidence. (a)~Diagnostic trial and cut evidence used to locate the phase switch. (b)~Selection of $\tau_Q$.}
  \label{fig:sensitivity}
\end{figure}

Thresholds are selected using the Qwen BFCL v4 Live $2{,}251$ items reported in Table~\ref{tab:all-results}. Excluding this model--task pair, the remaining nine pairs still reduce output tokens by $48.4\%$ and E2E time by $52.5\%$.

\paragraph{The reversal-count thresholds are $K_{\mathrm{phase}}=4$ and $K_{\mathrm{osc}}=8$.}
Figure~\ref{fig:sensitivity}a shows the two evidence lanes that delimit the phase switch. Trial conversion peaks at index~4 with $42.5\%$ of trials accepted and correct, then falls to $17.4\%$ at index~6, while the cut yield rises steadily with the reversal count and reaches $34.7\%$ at $K=8$ across $1{,}064$ trajectories. Although $K=11$ yields a nominally higher $37.7\%$, it relies on only $236$ trajectories. We fix $K_{\mathrm{phase}}=4$ as the phase-switch threshold, so that trials retire once the reversal count reaches four, and select $K_{\mathrm{osc}}=8$ from the cut lane, where the trajectory evidence is strongest with adequate support.

\paragraph{The net-displacement-ratio threshold is set to $\tau_Q=0.14$.}
Figure~\ref{fig:sensitivity}b plots accuracy and retained reasoning tokens against $\tau_Q$. We loosen the gate to enable more early stops while preserving accuracy. Raising the gate from $0.06$ to $0.14$ lifts trajectory stops from $699$ to $1{,}067$, a $52.6\%$ increase, and cuts retained reasoning tokens from $322.0$ to $240.9$, a $25.2\%$ reduction. At $0.14$, we observe the highest accuracy of $79.96\%$ with $1{,}800$ correct items. Its token count is also within $1.5\%$ of the lowest value in the sweep, so $0.14$ is close to the best on both accuracy and cost. Beyond $0.14$, tokens stay near their minimum while accuracy falls below the peak, yielding little additional saving. We therefore select $0.14$ to balance quality and token savings.

%% file: conclusion.tex
\section{Conclusion}
\label{sec:conclusion}

Early-exit methods usually decide from local statistics, yet whether reasoning is still making progress is a property of the whole trace. We propose GLOBE, an online, training-free algorithm that embeds each completed reasoning segment as a point on a unit sphere and reads the shape of the resulting semantic trajectory. Its trial phase stops the main stream on an accepted trial, and its trajectory phase stops it once the trajectory repeatedly reverses direction with little net progress. Across two models and five benchmarks, GLOBE cuts output tokens by 49.7\% and E2E time by 51.0\% while achieving a five-task average of 80.58\% against 80.04\% for Full CoT. The ablations show that every component matters: removing any one lowers the three-task average on both models, by up to 11.3 points. On GSM8K with Qwen, GLOBE stops 98.9\% of traces early and answers 96.3\% of them correctly. Combining confident trial answers with trajectory geometry keeps aggregate quality comparable to Full CoT at about half the output tokens and E2E time.

%% file: algorithm_appendix.tex
\section{Implementation details}
\label{app:implementation}

This appendix specifies the controller exactly as it runs: the execution
contract and accounting (\ref{app:implementation-details}), the segmenter and
geometry updates (\ref{app:segmentation}), degenerate cases
(\ref{app:formula-boundaries}), the per-increment control order with
Algorithm~\ref{alg:globe} (\ref{app:pseudocode}), the reference configuration
(\ref{app:reference-config}), the per-model and per-task details of the trial
branch (\ref{app:trial-details}), and the analysis-only spherical diagnostics
(\ref{app:intrinsic-sphere-diagnostics}).

\subsection{Execution Contract and Notation}
\label{app:implementation-details}

GLOBE uses the following notation and contracts for its online policy. The input $x$ comprises the user instruction, dialogue history, and tool schemas. The task-native template produces $c(x)=\operatorname{Tok}(\operatorname{Render}(x))$, and the autoregressive model defines $p_\theta$ over $\mathcal{V}$. The reasoning prefix is $r=(r_1,\ldots,r_T)$; $s_{\mathrm{close}}$ is the task-native thinking-close marker, such as \texttt{</think>}; and $y$ is the visible continuation containing the final answer or structured tool call. At each streaming step $t\le T$, policy $\pi$ observes $r_{\le t}$ and decides what to do.

The system maintains one continuous main reasoning stream and may launch bounded independent auxiliary trial-answer requests. Retained output tokens, discarded trial tokens, repeated prefix computation, request count, and total active E2E time are accounted for separately. Token-count claims count only retained tokens rather than collapsing distinct overheads into one quantity. Rejected trial text is discarded from its independent branch and never written to the main reasoning stream. An exit taken by the controller cancels the main stream, re-issues the retained prefix to the same serving endpoint so that its cached key-value state is reused, appends the task-native thinking-close marker, and generates the final answer or tool call through task-native formatting and parsing. A natural close preserves the original main stream.

At most $B$ trial answers may be launched per sample. Trial answers are disabled permanently when either $K_J$ reaches $K_{\mathrm{phase}}$ or the launched-trial count reaches $B$. Since the phase-switch check precedes candidate-signal evaluation on each increment, the increment causing the switch cannot launch another trial answer. Setting $B=0$ disables trials for a task without a matching scaffold while retaining trajectory-level geometry.

\subsection{Segmentation and Geometry Updates}
\label{app:segmentation}

The reference segmenter parses $R_t$ online, placing boundaries at sentences and newlines (\texttt{sent80}). Adjacent fragments are merged until each completed segment has at least $L_{\min}=80$ characters. The unfinished tail is not embedded. Newly completed segments are batched and truncated to at most 1200 characters before embedding; requests occur only when a segment completes.

The main stream maintains a latch for the thinking-close marker. When \texttt{</think>} is absent from the top-5, only the rising-edge latch resets; the growing prefix is not re-encoded. Re-encoding occurs only when an absent-to-present rising edge (Appendix~\ref{app:formula-boundaries}) satisfies the remaining candidate-signal conditions, avoiding quadratic recomputation while the marker is absent. The exact prefix of a \texttt{Wait} or \texttt{Actually} candidate signal ends immediately before the marker and excludes the marker itself; the exact prefix of a thinking-close rising edge contains all reasoning text generated through the current observation, because the delimiter has not been emitted.

\subsection{Formula Boundaries and Degenerate Cases}
\label{app:formula-boundaries}

Reversal statistics require at least three completed segments. With too few completed segments, the controller conservatively sets $K_J=0$ and $Q_J=1$, preventing a trajectory stop. The ratio is $Q_J=\delta_J/\ell_J$ for $\ell_J>0$ and $Q_J=1$ for $\ell_J=0$.

If $m_t=0$, no token can be scored. The trial answer is rejected because confidence is undefined. Non-finite values, dimension-incompatible vectors, or other embedding failures are handled fail-closed by permanently disabling the geometric observer for that sample. Any exception or error during a trial answer also rejects that trial.

The thinking-close marker has an absent-to-present rising edge when
\begin{equation}
s_{\mathrm{close}}\notin\operatorname{Top5}_{t-1}\quad\text{and}\quad s_{\mathrm{close}}\in\operatorname{Top5}_{t}.
\label{eq:rising-edge}
\end{equation}
The task-specific scaffold elicits a direct answer. Acceptance runs no separate format or schema check: the confidence score covers every token the trial returns, complete or not, and downstream task-native parsers provide the final format safety.

\subsection{Online Control Order and Policy Formalization}
\label{app:pseudocode}

For each streaming increment, the controller first commits completed segments and updates geometry, then applies the phase switch, evaluates the trajectory-stop condition, and finally evaluates at most one trial answer if still in the trial phase. Thus, the increment first reaching $K_{\mathrm{phase}}$ cannot launch another trial; a trajectory stop consumes no trial attempt; and rejected trial tokens never enter the retained prefix.

With $J(t)$ denoting the latest completed segment, the policy is
\begin{equation}
\pi(t)=\begin{cases}
\text{trajectory stop}, & \phi_t=\text{trajectory phase}\ \wedge\ K_{J(t)}\ge K_{\mathrm{osc}}\ \wedge\ Q_{J(t)}\le \tau_Q,\\
\text{trial-answer stop}, & \phi_t=\text{trial phase}\ \wedge\ \text{candidate signal holds}\ \wedge\ q_t>\tau_{\mathrm{conf}},\\
\text{natural close}, & \text{main stream emits }s_{\mathrm{close}},\\
\text{continue generation}, & \text{otherwise}.
\end{cases}
\label{eq:stopping-policy}
\end{equation}
Here $\phi_t\in\{\text{trial phase},\text{trajectory phase}\}$. Candidate-signal conditions include a 32-token cooldown, the limit $B$, discourse transitions, and a deduplicated thinking-close-marker rising edge.

\begin{algorithm}[t]
\caption{GLOBE online control loop (reference configuration)}
\label{alg:globe}
\begin{algorithmic}[1]
\STATE Initialize retained prefix $P\leftarrow c(x)$, $\phi\leftarrow\text{trial phase}$, $n_{\mathrm{trial}}\leftarrow0$, empty segment history, and Gram state.
\STATE Start one main reasoning stream (SSE) with top-5 log-probabilities.
\FOR{each new reasoning-text increment $\Delta R_t$}
  \STATE Append $\Delta R_t$ to $R_t$ and record newly completed segments.
  \IF{a segment completes}
    \STATE Request its embedding, normalize it, and incrementally update $S$, $K_J$, and $Q_J$.
  \ENDIF
  \IF{$K_J\ge4$}
    \STATE Permanently set $\phi\leftarrow\text{trajectory phase}$.
  \ENDIF
  \IF{$\phi=\text{trajectory phase}$ and $K_J\ge8$ and $Q_J\le0.14$}
    \STATE Cut at boundary $J(t)$; cancel the main stream, re-encode the prefix, append $s_{\mathrm{close}}$, generate $y$, and exit.
  \ENDIF
  \IF{$\phi=\text{trial phase}$ and a candidate signal holds and $n_{\mathrm{trial}}<B$}
    \STATE Re-encode the exact prefix; launch one serial trial of at most 20 tokens; increment $n_{\mathrm{trial}}$.
    \IF{$q_t>0.9$}
      \STATE Accept the snapshot; cancel the main stream, re-issue the retained prefix, append $s_{\mathrm{close}}$, generate $y$, and exit.
    \ELSE
      \STATE Discard the trial answer and retain the main stream.
    \ENDIF
  \ENDIF
\ENDFOR
\IF{the main stream emits $s_{\mathrm{close}}$}
  \STATE Close pending trial branches and side diagnostics, retain the reasoning, and decode $y$.
\ENDIF
\end{algorithmic}
\end{algorithm}

A natural close may occur in either phase. Actual emission of $s_{\mathrm{close}}$ retains the original stream without truncation or re-encoding, closes pending trial branches and side diagnostics, and generates $y$ with native task parameters. A top-5 rising edge only proposes a trial; it is not natural termination.

\subsection{Reference Configuration and Variant Boundaries}
\label{app:reference-config}

The reference thresholds $K_{\mathrm{phase}}=4$, $K_{\mathrm{osc}}=8$, and $\tau_Q=0.14$ are the reference operating point. Table~\ref{tab:contracts} lists the task-specific settings, budgets, and concurrency, including Character-LLM which uses no trials. The main stream uses \texttt{temperature=0}, \texttt{enable\_thinking=true}, \texttt{stream=true}, and \texttt{top\_logprobs=5}. BFCL v4 Live and Agent Memory run at concurrency 4 with a 16{,}384-token retained budget, and GSM8K and CSQA run at concurrency 8 with an 8{,}192-token budget; all methods use the same concurrency within each task. Every request allows two retries with a 0.5-second base interval and a 300-second timeout. These are serving controls, not stopping rules.

Each sample permits at most $B=10$ trials of at most $\tau_{\mathrm{trial}}=20$ tokens. The scaffold appended to the frozen prefix is \texttt{\textbackslash n**Final Answer**\textbackslash n\textbackslash boxed}; trial generation ends at the first closing brace or after 20 tokens, and the confidence of Eq.~(\ref{eq:confidence}) averages the chosen-token probability over every returned token, including the first one and a length-truncated tail, with $\tau_{\mathrm{conf}}=0.9$.

The only textual markers are \texttt{Wait} and \texttt{Actually}, matched with and without a leading space. The minimum segment length is $L_{\min}=80$, embedding input is capped at 1{,}200 characters, each completed segment permits one close-marker trial, and trials have a 32-token cooldown. Coincident textual and close-marker candidate signals are merged.

Trajectory representation uses \texttt{Qwen3-Embedding-8B} served on a separate vLLM instance. Without geometry, a sample cannot trajectory-stop. Empty trials, missing or misaligned log-probabilities, non-finite values, and zero-norm, non-finite, or dimension-incompatible embeddings are fail-closed: the trial is rejected or geometry is permanently disabled.

\subsection{Model- and Task-Specific Details of the Trial Branch}
\label{app:trial-details}

\paragraph{Thinking-close marker.}
The close marker is the tokenizer's own end-of-thinking token: \texttt{</think>}
(ID \texttt{248069}) for Qwen3.6-35B-A3B and \texttt{<channel|>} (ID
\texttt{101}) for Gemma-4-12B-it, whose reasoning is returned on a separate
reasoning channel. The top-5 rising edge of Eq.~(\ref{eq:rising-edge}) is
evaluated on this token, and the continuation appends the same token before
decoding the visible answer.

\paragraph{One scaffold for every task.}
GSM8K, CSQA, BFCL v4 Live, and Agent Memory share the boxed scaffold of
Appendix~\ref{app:reference-config}. The trial only has to reveal whether the
prefix already commits to an answer, so acceptance never parses the trial
text: a trial that starts a number, an option letter, or a function name inside
the box is scored the same way. The accepted continuation, not the trial, produces the answer
that the task parser reads, so the scaffold never changes the visible output
format; for tool calls the continuation is parsed by the serving stack's
native tool-call parser.

\paragraph{Role-play harness.}
Character-LLM answers are free-form in-character utterances, so the role-play
harness has no trial phase: it embeds completed segments, applies the trajectory
stop after a minimum number of completed segments, and answers from the
retained reasoning, with a 2{,}048-token generation budget and a 512-token
answer budget. Its thresholds are listed in Table~\ref{tab:contracts}.

\paragraph{Prefix reuse.}
Trials and continuations are sent to the endpoint that serves the main stream,
with prefix caching enabled, so the frozen prefix is not recomputed from
scratch. The repeated prefill that a trial or continuation would otherwise
cost is recorded per item.

\subsection{Intrinsic Spherical Diagnostics (Analysis Only)}
\label{app:intrinsic-sphere-diagnostics}

Intrinsic Riemannian spherical diagnostics are used only in offline shadow analysis. For $x,y\in\mathbb{S}^{d-1}$,
\begin{equation}
\theta(x,y)=\arccos\!\left(\operatorname{clip}(x^\top y,-1,1)\right).
\label{eq:geodesic}
\end{equation}
For non-degenerate, non-antipodal vectors,
\begin{equation}
\log_x(y)=\frac{\theta(x,y)}{\sin\theta(x,y)}\left(y-\cos\theta(x,y)x\right),
\label{eq:log-map}
\end{equation}
\begin{equation}
\operatorname{PT}_{x\to y}(w)=w-\frac{\langle w,y\rangle}{1+\langle x,y\rangle}(x+y),\qquad\langle w,x\rangle=0.
\label{eq:parallel-transport}
\end{equation}
Transporting $\log_{u_j}u_{j+1}$ to the tangent space at $u_{j+1}$ and comparing it with $\log_{u_{j+1}}u_{j+2}$ yields intrinsic turning angle, tangential-velocity change, and acceleration norm; degenerate and antipodal cases are recorded separately. These diagnostics do not affect control. The deployed policy uses only the Gram/chordal-displacement statistics and net-displacement ratio from Section~\ref{sec:method-sphere}.

%% file: theory_appendix.tex
\section{Mathematical characterization of the stopping statistics}\label{app:theory}

This appendix provides the detailed mathematical characterizations and formal derivations for the properties of the two stopping signals used by our method. Every statement presented in this section concerns the statistics exactly as they are deployed in the system (as described in Appendix~\ref{app:reference-config}). None of these formulations change the underlying controller logic, and none of them constitutes a statement about the correctness or truthfulness of a model's generated answer.

\subsection{The trial-answer score}\label{app:confidence-theory}

Let $\mathcal V$ be the discrete vocabulary of the language model, and let $\Delta(\mathcal V)$ denote the probability simplex over this vocabulary. The total variation distance between two probability distributions $P, Q \in \Delta(\mathcal V)$ is defined as $\mathrm{TV}(P,Q)=\tfrac12\|P-Q\|_1=\sup_{A\subseteq\mathcal V}|P(A)-Q(A)|$. For the next-token distribution at position $i$ along the trial we write $P_{t,i}$, and we abbreviate the probability of the emitted token as $p_{t,i}=P_{t,i}(\tilde y_i)$. Because the trial is decoded at temperature $0$, the emitted token is chosen greedily, $\tilde y_i\in\arg\max_{v\in\mathcal V} P_{t,i}(v)$, so that $p_{t,i}=\max_{v\in\mathcal V}P_{t,i}(v)=\|P_{t,i}\|_\infty$ is the largest probability of the distribution. If there are ties for the maximum probability, they are broken arbitrarily and do not affect any of the identities below. The deployed scorer then averages this confidence over every returned trial token. Note that an empty trial with $m_t=0$ is automatically rejected.

We now give the full derivation of the distance to point masses, which yields an equivalent total-variation characterization of the trial score in Eq.~(\ref{eq:confidence}). Fix any distribution $P\in\Delta(\mathcal V)$ and let $p_{\max}=\|P\|_\infty$. For any deterministic point-mass distribution $\delta_u$ concentrated at a single token $u$, the $\ell_1$ distance is evaluated by separating the element $u$ from the rest of the vocabulary:
\begin{align*}
\|P-\delta_u\|_1 &= |P(u)-1| + \sum_{v\neq u}|P(v)-0| \\
&= (1-P(u)) + \sum_{v\neq u}P(v) \\
&= (1-P(u)) + (1-P(u)) = 2(1-P(u)) .
\end{align*}
Dividing by $2$, the total variation distance to this point mass is exactly $\mathrm{TV}(P,\delta_u)=1-P(u)$. To find the closest point mass, we minimize over all tokens $u\in\mathcal V$. The minimum distance is achieved exactly at the modes of $P$, yielding $\min_u\mathrm{TV}(P,\delta_u)=1-p_{\max}$. This minimum distance coincides with the Bayes $0$--$1$ risk. If we make a randomised prediction $\hat Y\sim\pi$ for a true label $Y\sim P$, the probability of error is $1-\sum_v\pi(v)P(v)$. This error probability is bounded below by $1-p_{\max}\sum_v\pi(v)=1-p_{\max}$, with equality holding if and only if the prediction strategy $\pi$ is fully supported on the modes of $P$. Thus, the Bayes $0$--$1$ risk is exactly $1-p_{\max}$, and it is attained by any decision rule supported on the modes of $P$, including the greedy rule. This risk concerns predicting the model's own next token, not the correctness of the final answer. Furthermore, using the standard definition of the R\'enyi infinity entropy $H_\infty(P)=-\log\|P\|_\infty$, we have $p_{\max} = \exp(-H_\infty(P))$, meaning the minimum distance can also be written as $1-\exp(-H_\infty(P))$. Applying these identities at each trial position by setting $P=P_{t,i}$, the distance to the nearest deterministic distribution is $1-p_{t,i}$. Averaging this quantity over the $m_t$ trial positions yields exactly the complement of Eq.~(\ref{eq:confidence}).

The total variation distance can equivalently be written using its dual formulation. For any two distributions $P$ and $Q$, $\mathrm{TV}(P,Q)=\tfrac12\sup_{\|f\|_\infty\le1}|\mathbb E_Pf-\mathbb E_Qf|$, where the supremum is taken over all test functions bounded in magnitude by $1$. Consequently, the complement of the confidence score at a given position, $1-p_{t,i}$, is a dual-norm distance: it guarantees that no bounded test function evaluated on the next token can separate the predictive distribution $P_{t,i}$ from the deterministic law $\delta_{\tilde y_i}$ by more than $2(1-p_{t,i})$.

The single-step confidence can also be bounded via collision probabilities. For every distribution $P$, the properties of the $\ell_p$ norms dictate that $\|P\|_\infty^2\le\|P\|_2^2\le\|P\|_\infty$, which can be equivalently restated in terms of R\'enyi entropies as $H_\infty(P)\le H_2(P)\le 2H_\infty(P)$. To see why this holds algebraically, observe that $P(v) \le p_{\max}$ for all $v$, so multiplying by $P(v)$ and summing gives $\sum_vP(v)^2\le p_{\max}\sum_vP(v) = p_{\max}$. Conversely, since probabilities sum to $1$, the sum of squares is bounded below by the square of the maximum, $\sum_vP(v)^2\ge p_{\max}^2$. The quantity $\|P\|_2^2$ is exactly the probability that two independent draws from the distribution $P$ coincide. Therefore, at every trial position $i$, the probability that two independent samples agree lies in the interval $[p_{t,i}^2,p_{t,i}]$. Note that this is a single-pass, per-position quantity; it is not the multi-sample answer agreement measured by self-consistency decoding \citep{wang2023selfconsistency}.

Moving from single positions to the entire trial sequence, suppose $m=m_t\ge1$ and let the sequence $Y=(Y_1,\ldots,Y_m)$ be drawn position by position, with each coordinate $Y_i\sim P_{t,i}$ sampled from the next-token laws observed along the greedy trial. This product law is a surrogate that keeps the observed laws fixed; it is not the distribution of autoregressive sampling. Because the coordinates are drawn independently from these fixed laws, the probability that the entire sampled sequence matches the greedy trial exactly is the product of the individual probabilities: $\Pr(Y=\tilde y)=\prod_{i=1}^m p_{t,i}$. From this, we can establish bounds on the sequence-level error probability $\Pr(Y\neq\tilde y)=1-\prod_{i=1}^m p_{t,i}$ in terms of the average confidence $q_t$. On the left side, the arithmetic-mean geometric-mean (AM--GM) inequality implies that $\prod_{i=1}^m p_{t,i}\le (\frac1m\sum_{i=1}^m p_{t,i})^m = q_t^m$, which gives $1-q_t^{\,m}\le\Pr(Y\neq\tilde y)$. On the right side, a standard union bound over the positions at which $Y$ might depart from $\tilde y$ takes the form of the Weierstrass product inequality. Setting $a_i=1-p_{t,i}\in[0,1]$, we have $1-\prod_{i=1}^m(1-a_i)\le\sum_{i=1}^m a_i$, which implies $\Pr(Y\neq\tilde y)\le\sum_{i=1}^m(1-p_{t,i})=m\,(1-q_t)$. This establishes the two-sided inequality
\begin{equation}
1-q_t^{\,m}\;\le\;\Pr(Y\neq\tilde y)\;\le\;m\,(1-q_t).
\label{eq:app-seq-sandwich}
\end{equation}
The upper bound is practically informative only when it is strictly less than $1$, meaning $m_t(1-q_t)<1$. This occurs for short trials, such as generating boxed numerals. However, for a fully accepted $20$-token trial, the upper bound is informative only if $q_t>0.95$. Because of this looseness for longer sequences, the acceptance rule relies on interpreting the average confidence through the per-position identity of Eq.~(\ref{eq:confidence}), rather than through the sequence-level bound of Eq.~(\ref{eq:app-seq-sandwich}).

Finally, we address the algebraic choice of the aggregation function over the trial positions. The average confidence in Eq.~(\ref{eq:confidence}) is the arithmetic member $M_1$ of the generalized power means $M_\alpha=(\frac1m\sum_{i=1}^m p_{t,i}^\alpha)^{1/\alpha}$. The geometric mean $M_0=\exp(\frac1m\sum_{i=1}^m\log p_{t,i})$, used by DEER \citep{yang2025deer}, and the minimum $M_{-\infty}=\min_i p_{t,i}$ are also valid choices. By the monotonicity of power means, they satisfy the ordering $M_{-\infty}\le M_0\le M_1$. The arithmetic mean $M_1$ has a distinct robustness property: changing a single position's probability $p_{t,i}$ moves $M_1$ by at most $1/m_t$. This constitutes a bounded influence in the sense of \citet{hampel1974influence}. In contrast, one vanishing probability in the sequence sends the geometric mean $M_0$ entirely to $0$. Bounded influence cuts both ways. On the one hand, a single low-probability formatting token cannot unfairly veto a trial that is otherwise highly peaked. On the other hand, if the threshold is $\tau_{\mathrm{conf}}$, a single low-probability token cannot forcefully reject the trial either, provided the trial is long enough that $m_t>1/(1-\tau_{\mathrm{conf}})$.

\subsection{The unit-sphere trajectory statistics}\label{app:sphere-theory}

We now turn to the geometric properties of the trajectory on the unit sphere. Let the embedding space $\mathcal E=(\mathbb R^d,\langle\cdot,\cdot\rangle)$ be a real Hilbert space. The normalized segment embeddings form a sequence $u_1,\ldots,u_J\in\mathbb S^{d-1}$. Normalization removes the length of the raw embedding vectors and makes every statistic below invariant to their scale, in line with the motivation of Section~\ref{sec:method-sphere}. After normalisation, the geometric relationship between any two states $u_a$ and $u_b$ is entirely encapsulated by their inner product $S_{a,b}=\langle u_a,u_b\rangle$. The distance between them along the straight chord is $D_{a,b}=\|u_a-u_b\|$. By expanding the squared norm, we see that $D_{a,b}^2=\|u_a\|^2+\|u_b\|^2-2\langle u_a,u_b\rangle=2-2S_{a,b}$, so the chord length $D_{a,b}=\sqrt{2-2S_{a,b}}$ depends only on the angle $\theta_{a,b}$ between the vectors, and can be written as $2\sin(\theta_{a,b}/2)$. We denote the successive chordal displacements by $\xi_j=u_{j+1}-u_j$. Whenever angles between chords are evaluated, we assume that $\xi_j\neq0$, meaning that consecutive reasoning segments produce strictly distinct semantic embeddings.

The structural characteristics of the trajectory are fully determined by the inner products. In fact, if two families of points $\{u_a\}$ and $\{u'_a\}$ have the exact same Gram matrix $S$, they describe the identical shape. To prove this full Gram determinacy, define a linear map $T:\sum_a c_a u_a\mapsto\sum_a c_a u'_a$. For any linear combination, its squared norm is $\|\sum_a c_a u_a\|^2 = \sum_{a,b} c_a c_b S_{a,b}$, which equals $\|\sum_a c_a u'_a\|^2$ since the Gram matrices match. Thus, $T$ is well defined and isometric on $\operatorname{span}\{u_a\}$. By standard linear algebra, this map extends to a global linear isometry of the full space $\mathcal E$ mapping the orthogonal complement of $\operatorname{span}\{u_a\}$ isometrically onto that of $\operatorname{span}\{u'_a\}$, since both complements have the same dimension. Consequently, the Gram matrix $S$ determines the point sequence up to a global linear isometry, meaning our scalar statistics (which depend only on $S$) are invariant to orthogonal transformations of the embedding space and to strictly positive scaling of the raw unnormalized vectors $z_j\mapsto c_j z_j$. Furthermore, this formulation allows for an $O(d)$ streaming state update. When the new state $u_J$ arrives, computing the latest reversal metric $\kappa_{J-2}$ requires $S_{J-2,J-1}$ (computed and stored at the previous step), $S_{J-1,J}$, and $S_{J-2,J}$. The path length is updated as $\ell_J=\ell_{J-1}+\sqrt{2-2S_{J-1,J}}$, and the net displacement is updated as $\delta_J=\sqrt{2-2S_{1,J}}$. All three quantities require only evaluating three new inner products in $\mathbb R^d$. Thus, the controller's state memory can be bounded to just the previous two normalized embeddings $u_{J-2},u_{J-1}$, the initial embedding $u_1$, and the running scalar accumulators $K_{J-1}$, $\ell_{J-1}$, and $S_{J-2,J-1}$.

We can unpack the properties of the trajectory by examining the increments. Let $\Gamma$ be the Gram matrix of the increments themselves, with entries $\Gamma_{ij}=\langle \xi_i,\xi_j\rangle$ for $1\le i,j\le J-1$. By expanding the bilinear inner product of $\xi_i = u_{i+1}-u_i$ and $\xi_j = u_{j+1}-u_j$, we see that $\Gamma_{ij}$ is a mixed second difference of the base Gram matrix $S$: $\Gamma_{ij}=S_{i+1,j+1}-S_{i+1,j}-S_{i,j+1}+S_{i,j}$. The diagonal entries are simply the squared step lengths $\Gamma_{jj}=D_{j,j+1}^2$. The adjacent off-diagonal entries define the reversal signal $\kappa_j=\Gamma_{j,j+1}=\langle \xi_j,\xi_{j+1}\rangle$. By the polarization identity $2\langle a,b\rangle=\|a+b\|^2-\|a\|^2-\|b\|^2$ applied to $a=\xi_j$ and $b=\xi_{j+1}$, noting that $a+b=u_{j+2}-u_j$, we obtain $\kappa_j=\tfrac12(D_{j,j+2}^2-D_{j,j+1}^2-D_{j+1,j+2}^2)$. Substituting $D_{a,b}^2=2-2S_{a,b}$ into this polarization form algebraically recovers the expression in Eq.~(\ref{eq:gram-reversal}). A reversal occurs when $\kappa_j<0$, and $K_J$ counts the number of such reversals. The reversal count can also be read as $K_J = (J-2) - \#\{\text{persistent turns}\}$, where persistent turns are those with $\kappa_j\ge0$. The overall squared net displacement is the sum of all elements in $\Gamma$, since $u_J-u_1=\sum_{j=1}^{J-1} \xi_j$, meaning $\delta_J^2=\mathbf 1^\top\Gamma\mathbf 1$. Grouping these terms by their lag $|i-j|$ produces a clear displacement decomposition:
\begin{equation}
\delta_J^2=\mathbf 1^\top\Gamma\mathbf 1=\sum_{j=1}^{J-1}\|\xi_j\|^2+2\sum_{j=1}^{J-2}\kappa_j+2\sum_{h=2}^{J-2}\sum_{j=1}^{J-1-h}\langle \xi_j,\xi_{j+h}\rangle .
\label{eq:app-displacement-decomposition}
\end{equation}
This exact identity demonstrates the interaction between the net displacement and the local turns: every trajectory reversal ($\kappa_j<0$) contributes a negative lag-one term that actively reduces the squared net displacement, whereas persistent turns ($\kappa_j\ge0$) add positively to it.

To bound the net-displacement ratio $Q_J = \delta_J / \ell_J$ of Section~\ref{sec:method-sphere}, consider paths with $\ell_J>0$. First, the triangle inequality $\delta_J\le\ell_J$ gives $Q_J\le 1$. Second, the spatial diameter of the unit sphere limits the displacement to $\delta_J\le\|u_J\|+\|u_1\|=2$, yielding $Q_J\le 2/\ell_J$. Together, these provide the bound $Q_J \le \min\{1, 2/\ell_J\}$. As a direct consequence, the condition $Q_J\le \tau_Q$ is implied by the path length alone once $\ell_J\ge2/\tau_Q$, because then $Q_J\le2/\ell_J\le \tau_Q$. The converse does not hold: a short path whose endpoint returns close to its start satisfies $Q_J\le \tau_Q$ as well. In this sense $\tau_Q$ also acts as a path-length budget for long streams. Furthermore, $Q_J<1$ strictly as soon as the sequence visits three distinct points. The Hilbert space $\mathcal E$ is strictly convex. Equality in the triangle inequality $\delta_J=\sum_{j=1}^{J-1}\|\xi_j\|$ would force all nonzero step vectors $\xi_j$ to be positive scalar multiples of one another. This condition would compel every point $u_j$ to lie on a single ray emanating from $u_1$. However, a straight ray can intersect the spherical surface in at most two points. Thus, any trajectory with three or more distinct states must strictly satisfy $Q_J<1$.

We can relate the total turning of the trajectory to both the reversal count and the net displacement. Let the step directions be $e_j=\xi_j/\|\xi_j\|\in\mathbb S^{d-1}$, and let $\Theta_J=\sum_{j=1}^{J-2}\angle(\xi_j,\xi_{j+1})$ be the total turning angle. Since a reversal is defined as a turn where $\kappa_j<0$, equivalent to $\angle(\xi_j,\xi_{j+1})>\pi/2$, and all other turns have non-negative angles bounded by $\pi$, we immediately have $\Theta_J\ge(\pi/2) K_J$. For the second bound involving $Q_J$, connect the consecutive unit direction vectors $e_1,e_2,\ldots,e_{J-1}$ by minimizing great-circle arcs on $\mathbb S^{d-1}$. This forms a spherical path known as the tangent indicatrix of the chordal polygon, and its exact arc length is $\Theta_J$. If $\Theta_J\le\pi$, let $c$ be the midpoint of this path by arc length. Every individual direction vector $e_j$ must be within an arc length of $\Theta_J/2\le\pi/2$ from $c$ along the path. Because the geodesic distance is bounded by the path distance, the spherical distance from any $e_j$ to $c$ is at most $\Theta_J/2$. Taking the cosine of this distance bounds the inner product: $\langle e_j,c\rangle\ge\cos(\Theta_J/2)$. We then project the net displacement onto $c$:
\begin{equation}
\delta_J=\Bigl\|\sum_{j=1}^{J-1}\xi_j\Bigr\|\ge\Bigl\langle\sum_{j=1}^{J-1}\xi_j,c\Bigr\rangle=\sum_{j=1}^{J-1}\|\xi_j\|\langle e_j,c\rangle\ge\cos(\Theta_J/2)\sum_{j=1}^{J-1}\|\xi_j\|=\cos(\Theta_J/2)\,\ell_J .
\label{eq:app-indicatrix}
\end{equation}
Dividing by $\ell_J$ shows that $Q_J\ge\cos(\Theta_J/2)$. Consequently, a small ratio $Q_J\le \tau_Q$ forces the total turning to be large, $\Theta_J\ge 2\arccos \tau_Q$: if $\Theta_J\le\pi$, then $\cos(\Theta_J/2)\le Q_J\le \tau_Q$ and the bound follows because the cosine is decreasing on $[0,\pi/2]$; if $\Theta_J>\pi$, the bound holds trivially since $2\arccos \tau_Q\le\pi$ for $\tau_Q\ge0$. This is a discrete form of the classical geometric comparison between chord length and arc length for curves of bounded total curvature \citep{fenchel1929kruemmung,milnor1950curvature}.

Although related, the trajectory reversal count $K_J$ and the net-displacement ratio $Q_J$ do not mathematically determine each other. We prove this using two explicit constructions on the sub-sphere $\mathbb S^2\subset\mathbb S^{d-1}$ parameterized by $\sigma(\theta,\phi)=(\cos\phi\cos\theta,\cos\phi\sin\theta,\sin\phi)$. The first construction is a zig-zag with drift: let $u_j=\sigma\bigl((j-1)\alpha,(-1)^j\beta\bigr)$ with step parameters $\alpha=1/(J-1)$ and $\beta=0.6\alpha$. In this trajectory, each step advances by a steady $\alpha$ in longitude, but alternates in latitude with a total swing of $2\beta=1.2\alpha>\alpha$. For small steps the chord $\xi_j$ is approximately $(\alpha,\pm2\beta)$ in local longitude--latitude coordinates, so that $\langle \xi_j,\xi_{j+1}\rangle\approx\alpha^2-4\beta^2=-0.44\,\alpha^2<0$: the latitude oscillation dominates, consecutive chords form an obtuse angle, and every turn is a reversal, $K_J=J-2$. However, because the longitudinal drift is perfectly steady, the net displacement remains a fixed, large fraction of the path length. Specifically, setting $J=10$ gives $K_{10}=8=K_{\mathrm{osc}}$ and $Q_{10}=0.619>\tau_Q$. The reversal count condition is fully met, but no stop is issued. A direct computation verifies that for this construction, $K_J=J-2$ and $Q_J\ge0.605$ for all sequence lengths $3\le J\le2000$. The second construction is a loop without reversals: let $u_j=\sigma\bigl(2\pi(j-1)/(J-1),0\bigr)$ with $J-1\ge5$. This traverses the equator exactly once in equal angular steps of $\omega=2\pi/(J-1)<\pi/2$. The cosine of the step angle dictates $\kappa_j=2\cos\omega-\cos2\omega-1=2\cos\omega(1-\cos\omega)>0$, which means there are strictly zero reversals and $K_J=0$. However, because the path completes a perfect closed circle, the final position matches the first ($u_J=u_1$), yielding a net displacement of exactly zero, so $Q_J=0$. Here the ratio condition is satisfied, but no stop is issued because there are no reversals.

We can gain intuition for these statistics through several canonical reference models. These are strictly interpretive and do not constitute assumptions made by the controller. For the first model, consider equicorrelated streams where $S_{a,b}=\rho<1$ for all distinct $a\neq b$. By substitution, $\kappa_j=\rho-1<0$ everywhere, so reversals happen at every step and $K_J=J-2$. The path length evaluates to $\ell_J=(J-1)\sqrt{2-2\rho}$ and the displacement is identically $\delta_J=\sqrt{2-2\rho}$, which yields $Q_J=1/(J-1)$, remarkably independent of the correlation parameter $\rho$. If we deploy the reference limits $(K_{\mathrm{osc}},\tau_Q)=(8,0.14)$, such an equicorrelated stream first satisfies the trajectory-stop condition exactly at $J=10$, because $K_{10}=8$ and $Q_{10}=1/9<0.14$. In this view, a stream with no coherent temporal structure is stopped at its very first admissible segment, and each persistent turn encountered postpones the stop. For a second model, consider stationary streams where $S_{a,b}=\rho(|a-b|)$ for some positive-definite correlation function $\rho$ with $\rho(0)=1$. The reversal metric evaluates to $\kappa_j=2\rho(1)-\rho(2)-1=\gamma(2)-2\gamma(1)$, where $\gamma(h)=1-\rho(h)=\tfrac12D_{a,a+h}^2$ is the variogram of the process. A reversal thus corresponds to local sub-additivity of the variogram, $\gamma(2)<2\gamma(1)$. The critical boundary $\gamma(2)=2\gamma(1)$ is attained by sequences with orthogonal consecutive increments, which serves as the discrete counterpart to Kolmogorov's Wiener spiral in Hilbert space \citep{kolmogorov1940wiener}. Within stationary processes, mean-reverting streams such as the autoregressive $\rho(h)=e^{-\lambda h}$ will reverse at every single turn because $\kappa_j=-(1-e^{-\lambda})^2<0$. In stark contrast, a steady rotating stream described by $\rho(h)=\cos(\omega h)$ with angular step $0<\omega<\pi/2$ never reverses at all. On this interpretive reading, $K_J$ measures the accumulated elapsed history restricted only to non-persistent turns. Persistent turns—where the stream maintains its prior direction—act as the events that continually hold the trajectory stop back.

The net-displacement ratio $Q_J$ possesses a direct maximum-likelihood interpretation in directional statistics. Suppose we model the normalized directions of motion $e_j=\xi_j/\|\xi_j\|$ as draws from a von Mises--Fisher distribution on $\mathbb S^{d-1}$. The density is $f(e)\propto\exp(\beta\langle\mu,e\rangle)$, where $\mu$ is the mean direction and $\beta$ is the concentration parameter. We weight each observation by its spatial step size $\|\xi_j\|$, yielding an arc-length-weighted log-likelihood proportional to $\sum_{j=1}^{J-1}\|\xi_j\|\log f(e_j)$. Expanding this sum gives $\ell_J\log C_d(\beta)+\beta\langle\mu,\sum_{j=1}^{J-1} \xi_j\rangle$, which equals $\ell_J\log C_d(\beta)+\beta\langle\mu,u_J-u_1\rangle$, where $C_d(\beta)$ is the required normalizing constant. For any $\beta>0$ this is maximized over unit vectors $\mu$ by aligning $\mu$ with the net displacement, $\hat\mu=(u_J-u_1)/\delta_J$, which leaves $\ell_J\log C_d(\beta)+\beta\delta_J$. Since $C_d(\beta)=\beta^{d/2-1}/\bigl((2\pi)^{d/2}I_{d/2-1}(\beta)\bigr)$ and $I_\nu'(\beta)=I_{\nu+1}(\beta)+(\nu/\beta)I_\nu(\beta)$, one has $\frac{d}{d\beta}\log C_d(\beta)=-I_{d/2}(\beta)/I_{d/2-1}(\beta)$. Setting the derivative $-\ell_J\,I_{d/2}(\beta)/I_{d/2-1}(\beta)+\delta_J$ to zero shows that the fitted concentration $\hat\beta$ is characterized by the ratio of modified Bessel functions $A_d(\hat\beta)=I_{d/2}(\hat\beta)/I_{d/2-1}(\hat\beta)$ must exactly equal $Q_J$ \citep{mardia2000directional,banerjee2005vmf}. Since $A_d$ increases strictly monotonically from $0$ to $1$, observing $Q_J$ is mathematically equivalent to observing the fitted concentration of the directions of motion. Thresholding $Q_J$ is therefore thresholding this concentration. For an embedding space of dimension $d=4096$, the reference threshold $\tau_Q=0.14$ corresponds to a fitted concentration of $\hat\beta\approx585$.

Finally, we analyze the stability of these geometric quantities under small embedding perturbations, such as those caused by reduced-precision arithmetic or numerical batching in the embedding service. Let $z_j,z'_j\in\mathbb R^d\setminus\{0\}$ be two sets of raw unnormalized embeddings, which normalize to $u_j$ and $u'_j$ respectively. Let $\eta=\max_j\|u_j-u'_j\|$ be the maximum normalized perturbation. To bound $\eta$ in terms of the raw vectors, let $a=\|z_j\|$, $b=\|z'_j\|$, and $c=\langle z_j,z'_j\rangle/(ab)$. Expanding the norms yields $\|u_j-u'_j\|^2=2-2c$ and $\|z_j-z'_j\|^2=a^2+b^2-2abc$. The non-negativity of the quadratic form $4(a^2+b^2-2abc)-(2-2c)(a+b)^2=2(1+c)(a-b)^2\ge0$ implies that $(2-2c) \le 4(a^2+b^2-2abc)/(a+b)^2$. Taking the square root gives exactly the inner-product case of the Dunkl--Williams inequality \citep{dunkl1964norm}: $\|u_j-u'_j\|\le 2\|z_j-z'_j\|/(\|z_j\|+\|z'_j\|)$, which bounds the normalized drift uniformly for every index $j$. We can now bound the differences in the trajectory statistics. Let $\xi'_j=u'_{j+1}-u'_j$. Because $\|\xi'_j-\xi_j\|\le\|u'_{j+1}-u_{j+1}\|+\|u'_j-u_j\|\le2\eta$, and any unit-sphere chord has norm bounded by $2$, the difference in the reversal dot products is bounded by applying the triangle inequality:
\begin{align*}
|\kappa'_j-\kappa_j| &= |\langle \xi'_j,\xi'_{j+1}\rangle-\langle \xi_j,\xi_{j+1}\rangle| \\
&\le \|\xi'_j-\xi_j\|\,\|\xi'_{j+1}\| + \|\xi_j\|\,\|\xi'_{j+1}-\xi_{j+1}\| \\
&\le (2\eta)(2) + (2)(2\eta) = 8\eta .
\end{align*}
By summing the perturbations over the segments, the absolute path length is disturbed by at most $|\ell_J-\ell'_J|\le2(J-1)\eta$, and the net displacement is perturbed by no more than the endpoints' shifts, $|\delta_J-\delta'_J|\le\|u'_J-u_J\|+\|u'_1-u_1\|\le2\eta$. Thus we have the rigorous stability bounds
\begin{equation}
|\kappa_j-\kappa'_j|\le8\eta,\qquad|\ell_J-\ell'_J|\le2(J-1)\eta,\qquad|\delta_J-\delta'_J|\le2\eta .
\label{eq:app-stability}
\end{equation}
As a direct consequence for the deployed controller, the categorical reversal count $K'_J$ exactly matches $K_J$ whenever the margin strictly exceeds the noise bound, $\min_j|\kappa_j|>8\eta$. For the net-displacement ratio, provided the path is long enough such that $\ell_J>2(J-1)\eta$, the disturbed path length $\ell'_J$ satisfies $\ell'_J\ge\ell_J-2(J-1)\eta$. The difference in ratios expands as $Q'_J-Q_J=(\delta'_J-\delta_J)/\ell'_J+Q_J(\ell_J-\ell'_J)/\ell'_J$. Applying the absolute bounds yields $|Q'_J-Q_J|\le\bigl(2\eta+2(J-1)\eta\,Q_J\bigr)/\bigl(\ell_J-2(J-1)\eta\bigr)$. Therefore, the critical decision threshold $Q_J\le \tau_Q$ remains unchanged whenever the decision margin $|Q_J-\tau_Q|$ is strictly greater than this ratio bound.

%% file: experiment_contracts_appendix.tex
\section{Experimental details}
\label{app:experimental-details}

\subsection{Benchmarks, metrics, and task contracts}
\label{app:contracts}

\paragraph{CSQA.}
We use the 1{,}221 validation questions of
CommonsenseQA~\citep{talmor2019commonsenseqa}, each with five options. GLOBE
answers zero-shot with the option letter in a box, decoded at temperature 0,
and a fixed answer parser maps the visible answer to one of the five labels;
the metric is accuracy.

\paragraph{GSM8K.}
We use the 1{,}319 test problems of GSM8K~\citep{cobbe2021gsm8k} with a fixed
five-shot exemplar manifest, and score strict exact match of the final
\texttt{\#\#\#\#} answer.

\paragraph{BFCL v4 Live.}
Because the live split of BFCL v4~\citep{patil2025bfcl} is revised upstream,
we evaluate a frozen 2{,}251-item snapshot rather than following the live
revision. The snapshot contains 1{,}335 AST items---258 simple, 1{,}037
multiple, 16 parallel, and 24 parallel-multiple calls---and 916
relevance- and irrelevance-detection items, namely 16 relevance questions and
900 irrelevance questions, 25 of which this snapshot stores in the relevance
file. The upstream
live files as of 2026-09-26 regroup the same 2{,}251 items as 1{,}351
AST and 900 detection items, so absolute scores are comparable only within one
snapshot. Tool schemas are rendered by the model's chat template, calls are
parsed by the serving stack's native tool-call parser, and the score is the
item-weighted accuracy over all live items under the official AST and
detection checkers.

\paragraph{BFCL v4 Agent Memory.}
Each of the three memory backends (KV, Vector, Recursive Sum) has 192 tasks,
of which 37 are prerequisite conversations that populate the memory and 155
are scored. Every model step is one controlled request with a 16{,}384-token
retained budget. Table~\ref{tab:all-results} reports the equal-weight mean of
the three official backend accuracies, the sum of the three backends' active
wall times, and the official per-task output tokens averaged over the
backends.

\paragraph{Character-LLM.}
We use the 857 single-turn interview questions of
Character-LLM~\citep{shao2023character}, whose original evaluation rubric
scores five aspects on a 1--7 Likert scale. We use a four-aspect adaptation of
that rubric: an independent judge model scores memorization, personality,
values, and hallucination and omits stability. We report the normalized acting
score $\mathrm{NAS}=100\,(\bar s-1)/6$ for the mean aspect score $\bar s$;
NAS is our normalization of the mean judge score, not the metric defined in the
original paper. The judge is \texttt{cyankiwi/Qwen3.8-27B-AWQ-INT4} (4-bit AWQ
weights, group size 32, ModelScope revision \texttt{ffb29de}) served with vLLM
in non-thinking mode at temperature 0. All methods are judged by the same judge
on the same 857 questions.

\begin{table}[ht]
\centering
\caption{Task contracts and GLOBE settings. Budget is the retained-token budget
of a request; Conc.\ is the number of concurrent items, held fixed across all
methods within each task. GLOBE thresholds are
$(K_{\mathrm{phase}},K_{\mathrm{osc}},\tau_Q)$; Character-LLM has no trial
phase, and its first entry $n_{\min}$ is the number of completed segments
before the trajectory stop is armed.}
\label{tab:contracts}
\small
\setlength{\tabcolsep}{4pt}
\begin{tabular}{lrlrrcc}
\toprule
Benchmark & Items & Metric & Budget & Conc. & GLOBE (Qwen) & GLOBE (Gemma)\\
\midrule
CSQA & 1{,}221 & accuracy & 8K & 8 & $(4,8,0.14)$ & $(4,8,0.14)$\\
GSM8K & 1{,}319 & strict EM & 8K & 8 & $(4,8,0.14)$ & $(4,8,0.14)$\\
BFCL v4 Live & 2{,}251 & All-live acc. & 16K & 4 & $(4,8,0.14)$ & $(4,8,0.14)$\\
Agent Memory & $3\times155$ & agent acc. & 16K/step & 4 & $(4,8,0.14)$ & $(4,8,0.14)$\\
Character-LLM & 857 & NAS & 2K & 1 & $(n_{\min}{=}3,8,0.14)$ & $(n_{\min}{=}10,8,0.14)$\\
\bottomrule
\end{tabular}
\end{table}

\paragraph{Answer parsing and evaluation.}
Gemma and Qwen follow the same task-specific answer-parsing and evaluation protocol described above.

\subsection{Baselines}
\label{app:baselines}

All baselines use the same models, benchmark items, and vLLM serving stack as
GLOBE and follow their method-native contracts.
\emph{Nothink} disables the thinking channel through the chat template and
answers directly. \emph{Full CoT} leaves thinking unconstrained up to the
task's budget. \emph{DEER}~\citep{yang2025deer} issues a short boxed trial
answer at reflection markers such as \texttt{Wait} and exits once the trial's
confidence passes its threshold; it has neither a trajectory stop nor a phase
switch. \emph{ThinkBrake}~\citep{oh2025thinkbrake}
compares, at every sentence boundary, the log-probability of the most likely
continuation token with that of the thinking-close token, and closes thinking
once the margin falls below a task-level threshold $\tau$.
\emph{RCPD}~\citep{wei2026rcpd} tracks the rank of the thinking-close token at
sentence boundaries and force-closes thinking with its published
reasoning-dynamics rules. ThinkBrake and RCPD issue no trial requests.

\subsection{Cost accounting}
\label{app:cost-accounting}

\paragraph{Output tokens.}
For GSM8K, CSQA, BFCL v4 Live, and Character-LLM the token column counts the
retained output of the main stream---reasoning plus visible answer---re-encoded
with the model's own tokenizer from the stored strings, with the appended
close marker included for early-exited items. Trial requests, discarded trial
text, and prompts are excluded; the first two are reported separately in
Appendix~\ref{app:overhead}. Main-stream tokens decoded after the frozen prefix of an accepted trial are discarded with the stream and excluded as well; their elapsed time is covered by active E2E. Agent Memory uses the official per-task output
token counts.

\paragraph{Active E2E time.}
E2E is the wall time during which the run's requests are active, including the
main stream, trial requests, embedding requests, and continuations. For
Agent Memory it is the sum over the three backends, and for Character-LLM,
which runs one item at a time, the sum of per-item latencies. Overlapping generation and embedding work is counted once in wall time.

\paragraph{Aggregate reductions.}
The reductions quoted in Section~\ref{sec:experiments} compare each GLOBE cell
with the Full CoT cell of the same model and task and average the ten per-pair
reductions with equal weight, rather than pooling token or time volumes;
Appendix~\ref{app:reductions} lists the ten pairs and the mean excluding Qwen BFCL v4 Live.

\subsection{Serving and compute}
\label{app:compute}

All generation runs on NVIDIA A100 80GB PCIe GPUs with vLLM 0.22.1. Qwen3.6-35B-A3B is
served with tensor parallelism 4, a 200K context window, prefix caching, and
raw log-probabilities for the top-5 candidates; Gemma-4-12B-it is served with
tensor parallelism 2. The segment encoder
\texttt{Qwen3-Embedding-8B} runs as a separate vLLM embedding service on one
GPU. Trials and continuations go to the same generation endpoint as the main
stream so that its prefix cache is reused.

%% file: exit_appendix.tex
\section{Additional results}
\label{app:additional-results}

This appendix extends the three experimental questions of Section~\ref{sec:experiments}. Appendices~\ref{app:reductions} and~\ref{app:overhead} break the cost savings down by model--task pair and by request type; Appendices~\ref{app:early-stop-rates} and~\ref{app:two-model-stop} extend the stop-reliability analysis to CSQA and to Gemma; Appendices~\ref{app:csqa-ablation} and~\ref{app:character-ablation} extend the component ablation to CSQA and Character-LLM; Appendices~\ref{app:sweeps} and~\ref{app:encoder} report the net-displacement-ratio threshold sweep and the sensitivity to the segment encoder; and Appendix~\ref{app:signal-comparison} compares the candidate signals of the trial phase.

\subsection{Per-pair accuracy and cost changes}
\label{app:reductions}

\begin{table}[ht]
\centering
\caption{GLOBE against Full CoT on each model--task pair of
Table~\ref{tab:all-results}. $\Delta$Acc is the score difference in points (NAS for Character-LLM); token and E2E reductions are relative to Full CoT; the
speed-up is the ratio of the two E2E times. The ten-pair mean gives the figures quoted in the abstract; the final row excludes Qwen BFCL v4 Live.}
\label{tab:reductions}
\small
\setlength{\tabcolsep}{5pt}
\begin{tabular}{llrrrr}
\toprule
Model & Task & $\Delta$Acc (points) & Tok.\ reduction & E2E reduction & Speed-up\\
\midrule
\multirow{5}{*}{Qwen} & CSQA & $+1.7$ & $74.6\%$ & $76.0\%$ & $4.17\times$\\
 & Character-LLM & $+0.1$ & $31.8\%$ & $30.7\%$ & $1.44\times$\\
 & GSM8K & $-0.2$ & $55.5\%$ & $43.3\%$ & $1.76\times$\\
 & BFCL v4 Live & $-1.3$ & $61.4\%$ & $37.2\%$ & $1.59\times$\\
 & Agent Memory & $+3.0$ & $78.6\%$ & $64.9\%$ & $2.85\times$\\
\midrule
\multirow{5}{*}{Gemma} & CSQA & $-1.0$ & $64.5\%$ & $59.4\%$ & $2.46\times$\\
 & Character-LLM & $+0.9$ & $7.7\%$ & $75.0\%$ & $3.99\times$\\
 & GSM8K & $+0.8$ & $48.1\%$ & $43.0\%$ & $1.75\times$\\
 & BFCL v4 Live & $+3.5$ & $49.1\%$ & $52.5\%$ & $2.11\times$\\
 & Agent Memory & $-2.1$ & $26.0\%$ & $27.7\%$ & $1.38\times$\\
\midrule
\multicolumn{2}{l}{Mean over the ten pairs} & $+0.54$ & $49.7\%$ & $51.0\%$ & \\
\multicolumn{2}{l}{Mean excluding Qwen BFCL v4 Live} & $+0.74$ & $48.4\%$ & $52.5\%$ & \\
\bottomrule
\end{tabular}
\end{table}

Table~\ref{tab:reductions} shows that the savings of Section~\ref{sec:main-results}
are not carried by a few favorable pairs. GLOBE shortens all ten pairs, with
token reductions from $7.7\%$ to $78.6\%$ and E2E reductions from $27.7\%$ to
$76.0\%$; the E2E speed-up is at least $1.38\times$ everywhere and reaches
$4.17\times$ on Qwen CSQA. Accuracy moves
by at most $3.5$ points: six of the ten pairs gain, led by Gemma BFCL v4 Live at $+3.5$ and Qwen Agent Memory at $+3.0$ points, and no pair loses more than $2.1$ points. The largest token savings fall on the tasks where Full CoT writes the
longest traces, namely Agent Memory and CSQA on Qwen, where the trajectory stop has
the most repetition to remove.

\subsection{Controller overhead}
\label{app:overhead}

\begin{table}[ht]
\centering
\caption{What the controller costs, per item, for the Qwen runs behind Figure~\ref{fig:stop}b. Requests count the main stream, trials, and the continuation. The embedding
share is the embedding time divided by the item's wall time; because the
observer runs alongside the main stream, it bounds the latency the observer can
add.}
\label{tab:overhead}
\small
\setlength{\tabcolsep}{6pt}
\begin{tabular}{lrrr}
\toprule
 & CSQA & GSM8K & BFCL v4 Live\\
\midrule
\midrule
Trials launched per item & $0.05$ & $0.92$ & $1.36$\\
Trial acceptance rate (\%) & $41.3$ & $24.2$ & $20.4$\\
Discarded trial tokens per item & $0.2$ & $4.1$ & $20.0$\\
Requests per item & $2.04$ & $2.91$ & $3.06$\\
\midrule
Embedding requests per item & $4.82$ & $5.74$ & $2.80$\\
Embedding time share of item wall time (\%) & $13.6$ & $10.1$ & $10.5$\\
\bottomrule
\end{tabular}
\end{table}

\paragraph{Controller overhead in wall time.}
Trials are rare and short: at most $1.36$ per item, with at most $20$ discarded
trial tokens per item, which is $0.04\%$, $0.5\%$, and $5.4\%$ of the retained
output on CSQA, GSM8K, and BFCL v4 Live. An item issues two to three requests in
total, and each trial or continuation re-uses the prefix cache of the main
stream, as detailed in Appendix~\ref{app:trial-details}. Embedding requests take at most
$13.6\%$ of the per-item wall time and overlap with generation, so the E2E
reductions of Table~\ref{tab:reductions} already include the time overhead of trials, continuations, and the geometric observer.

\paragraph{The phase switch prunes trials.} The one-way phase switch also bounds the trial budget. Relative to the full-GLOBE run of the same sweep, allowing trials in the trajectory phase on BFCL v4 Live with Qwen raises the number of trials launched over the split by $54\%$, from $3{,}108$ to $4{,}781$, and the active E2E time by $4.5\%$, from $3{,}197$ to $3{,}342$ seconds, while accuracy changes by less than $0.5\%$. Retiring these trials therefore saves requests.

\subsection{Stop coverage versus stop quality}
\label{app:early-stop-rates}

Coverage alone cannot rank early-exit controllers, because stopping every trace would be reported as full coverage. Table~\ref{tab:early-stop-rates} therefore pairs each controller's stop rate with the accuracy among exactly those stopped items, and with their product, the share of all items that are stopped early and answered correctly, which is the green segment of Figure~\ref{fig:stop}b. The stop rate is the share of items that the controller itself stops, so a natural close never counts as a stop. Besides the three tasks of Figure~\ref{fig:stop}b, the table adds CSQA.

\begin{table}[ht]
\centering
\caption{Stop coverage versus stop quality on Qwen from separate analysis runs. Stop: stopped items over
$N$ (count in parentheses). Post-stop: accuracy among the stopped items. Yield:
stopped and correct over $N$. $N$ is $1{,}221$ (CSQA), $1{,}319$ (GSM8K),
$2{,}251$ (BFCL v4 Live), and $465$ (Agent Memory). On Agent Memory an item
counts as stopped when the request that ends it is stopped. Bold: best yield per task.}
\label{tab:early-stop-rates}
\small
\setlength{\tabcolsep}{4pt}
\begin{tabular}{llrrr}
\toprule
Task & Method & Stop (\%) & Post-stop (\%) & Yield (\%)\\
\midrule
\multirow{4}{*}{CSQA} & DEER & $58.9$ (719) & $77.7$ & $45.8$\\
 & ThinkBrake & $69.9$ (853) & $87.6$ & $61.2$\\
 & RCPD & $95.7$ (1{,}169) & $85.5$ & $81.9$\\
 & GLOBE (ours) & $98.6$ (1{,}204) & $86.2$ & $\mathbf{85.0}$\\
\midrule
\multirow{4}{*}{GSM8K} & DEER & $65.4$ (862) & $95.4$ & $62.3$\\
 & ThinkBrake & $0.0$ (0) & -- & $0.0$\\
 & RCPD & $92.6$ (1{,}222) & $96.6$ & $89.5$\\
 & GLOBE (ours) & $98.9$ (1{,}304) & $96.3$ & $\mathbf{95.2}$\\
\midrule
\multirow{4}{*}{BFCL v4 Live} & DEER & $49.4$ (1{,}113) & $78.8$ & $39.0$\\
 & ThinkBrake & $5.8$ (130) & $66.9$ & $3.9$\\
 & RCPD & $77.7$ (1{,}749) & $59.8$ & $46.5$\\
 & GLOBE (ours) & $70.1$ (1{,}577) & $76.0$ & $\mathbf{53.3}$\\
\midrule
\multirow{3}{*}{Agent Memory} & DEER & $36.8$ (171) & $50.3$ & $18.5$\\
 & RCPD & $30.5$ (142) & $64.1$ & $19.6$\\
 & GLOBE (ours) & $46.9$ (218) & $59.2$ & $\mathbf{27.7}$\\
\bottomrule
\end{tabular}
\end{table}

\paragraph{GLOBE has the highest yield on all four tasks.}
GLOBE converts the largest share of items into correct early exits on every
task: $85.0\%$ on CSQA, $95.2\%$ on GSM8K, $53.3\%$ on BFCL v4 Live, and
$27.7\%$ on Agent Memory, ahead of the best baseline by $3.1$, $5.7$, $6.8$, and $8.1$ points. On CSQA and GSM8K it combines the highest coverage at $98.6\%$ and
$98.9\%$ with diagnostic post-stop accuracies of $86.2\%$ and $96.3\%$.

\paragraph{The baselines fail on different axes.}
ThinkBrake is the most accurate on the items it stops on CSQA at $87.6\%$ but
reaches only $69.9\%$ coverage there, never triggers on GSM8K, and stops $5.8\%$ of
BFCL v4 Live. DEER needs a reflection marker followed by a confident trial, so
its coverage stays between $36.8\%$ and $65.4\%$ and its post-stop accuracy on
CSQA is $77.7\%$, $8.5\%$ below GLOBE's. RCPD covers the most items after
GLOBE on CSQA and GSM8K, but on BFCL v4 Live the $77.7\%$ of items it closes are
only $59.8\%$ correct, against $76.0\%$ for GLOBE: coverage bought by closing
the stream before the call structure settles is paid for in accuracy. On Agent
Memory RCPD's stopped items are slightly more accurate at $64.1\%$ against
$59.2\%$, but it stops only $30.5\%$ of the items, so its yield is $19.6\%$
against $27.7\%$. Read as a pair, no baseline reaches GLOBE's coverage at GLOBE's
post-stop accuracy on any task.

\subsection{Stop reliability on both models}
\label{app:two-model-stop}

\begin{table}[ht]
\centering
\caption{Harm and rescue of GLOBE early stops against Full CoT from separate analysis runs, joined item by
item as in Figure~\ref{fig:stop}a. Harm: share of Full CoT-correct items that
GLOBE stops early and answers wrong. Rescue: share of Full CoT-wrong items that
GLOBE stops early and answers right.}
\label{tab:two-model-stop}
\small
\setlength{\tabcolsep}{6pt}
\begin{tabular}{llrr}
\toprule
Model & Task & Harm (\%) & Rescue (\%)\\
\midrule
\multirow{3}{*}{Qwen} & GSM8K & $0.94$ & $29.8$\\
 & BFCL v4 Live & $4.17$ & $9.56$\\
 & Agent Memory & $10.8$ & $24.9$\\
\midrule
\multirow{2}{*}{Gemma} & GSM8K & $3.14$ & $40.0$\\
 & BFCL v4 Live & $4.14$ & $9.75$\\
\bottomrule
\end{tabular}
\end{table}

Table~\ref{tab:two-model-stop} repeats the harm--rescue analysis of Figure~\ref{fig:stop}a on Gemma. The pattern transfers across model families. On GSM8K, Gemma's early stops rescue $40.0\%$ of the items that Full CoT gets wrong while harming $3.14\%$ of those it gets right; on BFCL v4 Live its two rates, $4.14\%$ and $9.75\%$, are within $0.2\%$ of Qwen's. On both models, successful early answers are observed even on items that Full CoT answers incorrectly.

\subsection{Gemma GSM8K without the trajectory stop}
\label{app:gemma-gsm-protocol}

Table~\ref{tab:remove-one} reports $62.2\%$ for Gemma on GSM8K without the trajectory stop. This run keeps the phase switch and the trial branch, and its items fall into three mutually exclusive outcomes:
\begin{center}
\begin{tabular}{lrr}
\toprule
Outcome & Items & Correct\\
\midrule
Trial stop & 111 & 92\\
Natural close without budget exhaustion & 737 & 728\\
No trial stop, budget exhausted & 471 & 0\\
\midrule
Total & 1{,}319 & 820\\
\bottomrule
\end{tabular}
\end{center}
All $471$ items that reach the $8{,}192$-token budget without a trial stop are answered incorrectly, whereas $728$ of the $737$ natural closes are correct. The loss therefore comes from traces that keep reasoning until the budget runs out, which is the behavior the trajectory stop is designed to cut.

\subsection{CSQA remove-one ablation}
\label{app:csqa-ablation}

\begin{table}[ht]
\centering
\caption{CSQA remove-one ablation ($N=1{,}221$). Every variant is a full run
compared item by item with the full-GLOBE run of the same sweep. $\Delta$:
change in correct items; $p$: exact McNemar test; Stop: share of items stopped
early; Tok: retained reasoning tokens relative to the full run.}
\label{tab:csqa-ablation}
\small
\setlength{\tabcolsep}{3.5pt}
\begin{tabular}{l|rrrr|rrrr}
\toprule
 & \multicolumn{4}{c|}{\textbf{Qwen}} & \multicolumn{4}{c}{\textbf{Gemma}}\\
Variant & $\Delta$ & $p$ & Stop & Tok & $\Delta$ & $p$ & Stop & Tok\\
\midrule
GLOBE (full) & -- & -- & $97.0\%$ & $1.00\times$ & -- & -- & $76.7\%$ & $1.00\times$\\
\quad w/o trial answers & $-7$ & $0.38$ & $97.0\%$ & $1.01\times$ & $+1$ & $1.00$ & $77.0\%$ & $1.07\times$\\
\quad w/o trajectory stop & $\mathbf{-127}$ & $2{\times}10^{-24}$ & $1.9\%$ & $6.84\times$ & $\mathbf{-107}$ & $5{\times}10^{-11}$ & $8.0\%$ & $7.71\times$\\
\quad w/o phase switch & $-5$ & $0.60$ & $96.8\%$ & $0.96\times$ & $+1$ & $1.00$ & $77.5\%$ & $0.95\times$\\
\quad w/o $\tau_Q$ gate & $+7$ & $0.41$ & $98.8\%$ & $0.85\times$ & $-16$ & $0.12$ & $84.0\%$ & $0.93\times$\\
\bottomrule
\end{tabular}
\end{table}

The trajectory stop is the load-bearing component on CSQA as well. Removing it costs $127$ correct items on Qwen and $107$ on Gemma, that is, $10.4$ and $8.8$ points, with $p=2\times10^{-24}$ and $5\times10^{-11}$. It also collapses the early-stop share from $97.0\%$ and $76.7\%$ to $1.9\%$ and $8.0\%$ and multiplies the retained reasoning by $6.8$ and $7.7$, consistent with a model that keeps re-examining options it has already ruled out and eventually talks itself out of the right one. The other three removals stay within noise on this task with $|\Delta|\le16$ items and $p\ge0.12$, and move the token count by at most $15\%$, so the trial phase and the $\tau_Q$ gate act as safeguards rather than as the main exit.

\subsection{Character-LLM remove-one ablation}
\label{app:character-ablation}

\begin{table}[ht]
\centering
\caption{Character-LLM remove-one ablation ($N=857$ single-turn interviews).
Each variant is a full run of the role-play harness compared item by item with
the full-GLOBE run of the same sweep. $\Delta$NAS in points with a paired bootstrap $95\%$
interval; $p$: exact Wilcoxon signed-rank test; Tok: retained reasoning tokens
relative to the full run. The role-play harness has no trial phase
(Appendix~\ref{app:trial-details}), so there is no \emph{w/o trial answers}
row.}
\label{tab:character-ablation}
\small
\setlength{\tabcolsep}{3pt}
\begin{tabular}{l|rrr|rrr}
\toprule
 & \multicolumn{3}{c|}{\textbf{Qwen}} & \multicolumn{3}{c}{\textbf{Gemma}}\\
Variant & $\Delta$NAS [95\% CI] & $p$ & Tok & $\Delta$NAS [95\% CI] & $p$ & Tok\\
\midrule
\quad w/o trajectory stop & $+1.49$ [$+0.32$, $+2.66$] & $0.08$ & $2.78\times$ & $+1.83$ [$+0.56$, $+3.07$] & $6{\times}10^{-10}$ & $1.43\times$\\
\quad w/o phase switch & $+0.16$ [$-1.03$, $+1.37$] & $0.97$ & $1.00\times$ & $-0.02$ [$-1.20$, $+1.18$] & $0.80$ & $1.01\times$\\
\quad w/o $\tau_Q$ gate & $-0.33$ [$-1.58$, $+0.92$] & $0.69$ & $0.89\times$ & $-2.05$ [$-3.25$, $-0.86$] & $8.5{\times}10^{-4}$ & $0.88\times$\\
\bottomrule
\end{tabular}
\end{table}

On role-play the trajectory stop acts as a cost control rather than an accuracy
control. Disabling it raises NAS by at most $1.8$ points while the retained reasoning
grows $2.78\times$ on Qwen and $1.43\times$ on Gemma, and on Gemma $14$ of the
$857$ interviews then fail to return an answer inside the $2{,}048$-token budget,
so the stop is also what keeps role-play generation bounded. The $\tau_Q$ gate
is the component that protects answer quality here: removing it lowers NAS by $2.05$ points on Gemma with $p=8.5\times10^{-4}$, consistent with the veto holding
back stops while the character's answer is still being developed. Because the role-play harness has no trial phase, the phase switch here only acts as the warm-up gate that keeps the trajectory stop from arming before $n_{\min}$ completed segments. The first accepted stops already fall after this gate, so removing the switch is a negative control: NAS changes by at most $0.16\%$ with $p\ge0.80$, and the token count stays at $1.00\times$ and $1.01\times$.
No removal yields a significant NAS gain without raising the cost.

\subsection{Operating-point sweeps on BFCL v4 Live}
\label{app:sweeps}

Table~\ref{tab:tauq-sweep} reports the net-displacement-ratio threshold sweep on all $N=2{,}251$ BFCL v4 Live items with Qwen.

\begin{table}[ht]
\centering
\caption{$\tau_Q$ sweep with $K_{\mathrm{phase}}=4$ and $K_{\mathrm{osc}}=8$.
Tok: retained reasoning tokens per item. Traj.: trajectory stops. Bold: the selected setting.}
\label{tab:tauq-sweep}
\small
\setlength{\tabcolsep}{5pt}
\begin{tabular}{rrrrrr}
\toprule
$\tau_Q$ & Correct & Acc.\ (\%) & Tok & E2E (s) & Traj.\\
\midrule
$0.06$ & 1{,}798 & $79.88$ & $322.0$ & 3{,}332 & 699\\
$0.08$ & 1{,}786 & $79.34$ & $285.6$ & 3{,}253 & 835\\
$0.10$ & 1{,}790 & $79.52$ & $270.8$ & 3{,}325 & 966\\
$0.12$ & 1{,}783 & $79.21$ & $252.9$ & 3{,}105 & 1{,}019\\
$\mathbf{0.14}$ & $\mathbf{1{,}800}$ & $\mathbf{79.96}$ & $\mathbf{240.9}$ & $\mathbf{3{,}071}$ & $\mathbf{1{,}067}$\\
$0.16$ & 1{,}785 & $79.30$ & $241.8$ & 2{,}878 & 1{,}070\\
$0.18$ & 1{,}770 & $78.63$ & $244.9$ & 3{,}027 & 1{,}030\\
$0.20$ & 1{,}777 & $78.94$ & $238.1$ & 3{,}010 & 1{,}040\\
$0.22$ & 1{,}777 & $78.94$ & $246.7$ & 3{,}112 & 1{,}050\\
$0.24$ & 1{,}783 & $79.21$ & $253.8$ & 3{,}191 & 1{,}060\\
$0.26$ & 1{,}778 & $78.99$ & $244.7$ & 3{,}086 & 1{,}061\\
$0.28$ & 1{,}771 & $78.68$ & $240.7$ & 2{,}987 & 1{,}081\\
$0.30$ & 1{,}785 & $79.30$ & $237.5$ & 2{,}994 & 1{,}047\\
\bottomrule
\end{tabular}
\end{table}

\paragraph{$\tau_Q$.} Loosening the gate from $0.06$ to $0.14$ raises the number of trajectory stops by $368$, from $699$ to $1{,}067$, and cuts the retained reasoning by $25\%$ and the E2E time by $8\%$. At $0.14$, we observe the highest accuracy in this threshold sweep. Beyond $0.14$ the stop count saturates at $1{,}030$--$1{,}081$ and the token count at $237$--$254$, yielding at most modest extra token savings alongside $15$--$30$ fewer correct items. We select $0.14$ as the knee at which the saving saturates.

The default setting $(K_{\mathrm{phase}}, K_{\mathrm{osc}}, \tau_Q) = (4, 8, 0.14)$, already reported in Section~\ref{sec:evidence-horizon-selection}, yields $1{,}800$ correct answers out of $2{,}251$, an accuracy of 79.96\%, together with 240.9 retained reasoning tokens per item and $1{,}067$ trajectory stops.

\subsection{Sensitivity to the segment encoder}
\label{app:encoder}

GLOBE embeds every completed segment with \texttt{Qwen3-Embedding-8B}. To test whether the trajectory stop depends on this particular encoder, we run GLOBE on Qwen with the reference configuration, which uses \texttt{sent80} segmentation, $(K_{\mathrm{phase}},K_{\mathrm{osc}},\tau_Q)=(4,8,0.14)$, and the trial settings of Appendix~\ref{app:reference-config}. We replace only the encoder, by \texttt{Qwen3-Embedding-4B} or \texttt{Qwen3-Embedding-0.6B}, each served as its own vLLM embedding instance, on BFCL v4 Live and on GSM8K.

\begin{table}[ht]
\centering
\caption{Sensitivity to the segment encoder (Qwen, reference configuration).
The 8B rows use the default encoder; on BFCL v4 Live this is the run of
Table~\ref{tab:all-results}. Tok: retained reasoning tokens per item. Traj.:
items closed by the trajectory stop. Embed: embedding time per item.}
\label{tab:encoder}
\small
\setlength{\tabcolsep}{4pt}
\begin{tabular}{llrrrrrr}
\toprule
Task & Encoder & Dim. & Acc.\ (\%) & Tok & E2E (s) & Traj. & Embed (s)\\
\midrule
\multirow{3}{*}{BFCL v4 Live} & 8B & 4{,}096 & $79.96$ & $240.9$ & 3{,}071 & 1{,}067 & $0.54$\\
 & 4B & 2{,}560 & $79.61$ & $236.2$ & 3{,}081 & 1{,}063 & $0.43$\\
 & 0.6B & 1{,}024 & $79.12$ & $242.5$ & 3{,}177 & 1{,}063 & $0.37$\\
\midrule
\multirow{3}{*}{GSM8K} & 8B & 4{,}096 & $95.91$ & $358.3$ & 1{,}751 & 1{,}032 & $0.95$\\
 & 4B & 2{,}560 & $96.36$ & $357.3$ & 1{,}893 & 1{,}030 & $0.78$\\
 & 0.6B & 1{,}024 & $95.91$ & $362.7$ & 1{,}762 & 1{,}035 & $0.58$\\
\bottomrule
\end{tabular}
\end{table}

\paragraph{Accuracy and stop counts are similar with a 13-fold smaller encoder.} On BFCL v4 Live the trajectory stop closes $1{,}063$ items with either smaller encoder against $1{,}067$ with the default one, and on GSM8K $1{,}030$ and $1{,}035$ items against $1{,}032$. The retained reasoning stays within $2\%$ of the default encoder on both tasks, at $236$--$243$ tokens on BFCL v4 Live and $357$--$363$ on GSM8K. Accuracy changes by at most $0.84\%$, and no change is significant, with all $p>0.05$; on GSM8K the 4B encoder answers six more items than the default encoder and the 0.6B encoder the same number. Active E2E time does not vary systematically with encoder size. The robustness can be plausibly explained by the construction of Section~\ref{sec:method-sphere}: the stopping statistics depend only on the angles between segment embeddings, and the reversal test only on the sign of $\kappa_j$, so the decisions can be insensitive to encoder changes if they preserve the coarse geometry of the trace. A smaller encoder also shortens the embedding time per item by up to $39\%$, from $0.95$ to $0.58$ seconds on GSM8K, which lowers the footprint of the geometric observer measured in Appendix~\ref{app:overhead}.

%% file: mechanism_audit_appendix.tex
\subsection{Which candidate signal authorizes an early exit}
\label{app:signal-comparison}
\suppressfloats[t]
\begin{figure}[H]
  \centering
\includegraphics[width=0.92\textwidth]{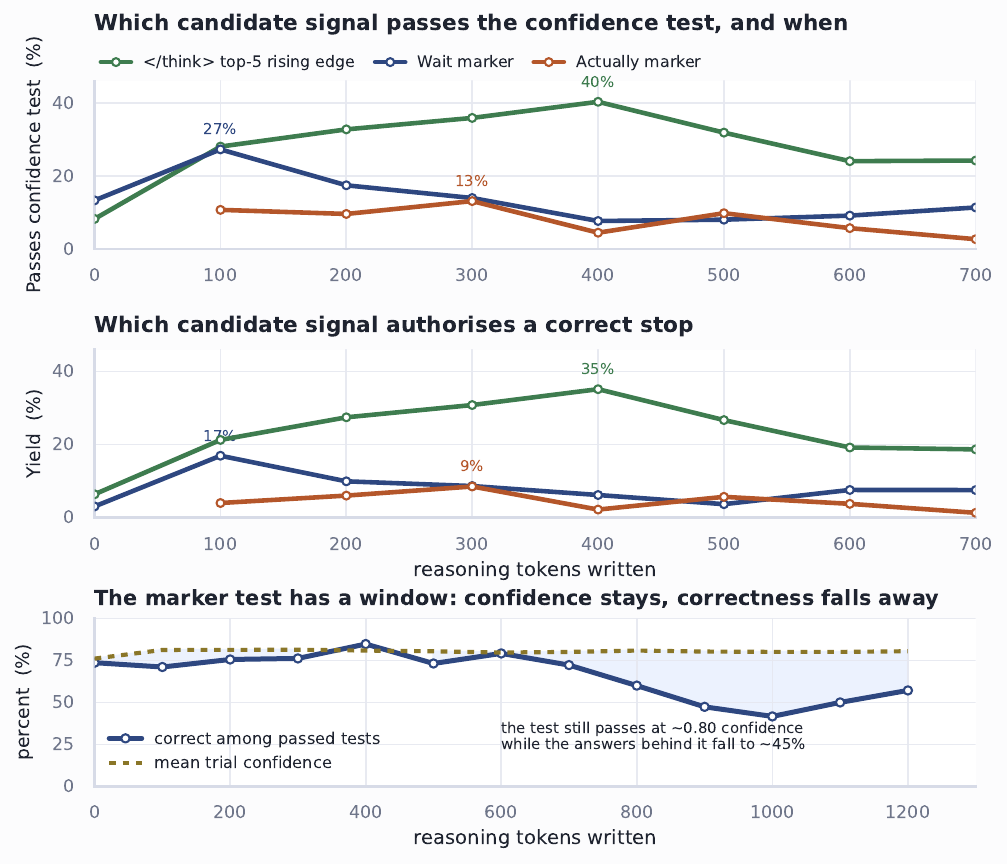}
  \caption{Signal-specific analysis. Top: the share of candidate
  signals whose short trial passes the confidence test, split by signal type.
  Middle: the share that passes the test and is correct. The thinking-close signal
  keeps improving until $400$ tokens ($40\%$ gate, $35\%$ yield), while the two
  textual markers never exceed $27\%$ and $17\%$; \texttt{Wait} peaks at
  $100$--$200$ tokens and decays from there, and \texttt{Actually} is the weakest
  of the three throughout. Bottom, over a longer axis: the marker lane's window.
  The share of passed confidence tests whose answer is correct falls from $78\%$ to $45\%$
  beyond $800$ tokens while the mean trial confidence stays flat at about $0.80$. Bins are $100$ tokens wide and hold at least $25$ events.}
  \label{fig:signal-comparison}
\end{figure}

For Figure~\ref{fig:two-triggers}a, trial events are binned by the completed-segment index $J-2$, with each bin placed at its median reasoning-token count.

The marker lane of Figure~\ref{fig:two-triggers} pools three candidate
signals, and they are not interchangeable. Pooled over the deployment-conditioned
population the thinking-close signal reaches a gate rate of $21.1\%$ and a yield
of $16.8\%$, \texttt{Wait} reaches $14.7\%$ and $8.9\%$, and \texttt{Actually}
reaches $7.9\%$ and $4.5\%$. Figure~\ref{fig:signal-comparison} shows why pooling
them hides the structure: the close signal's gate rate and yield climb to $40\%$
and $35\%$ at $400$ tokens, whereas \texttt{Wait} peaks at $27\%$ and $17\%$
around $100$ tokens and then decays, and \texttt{Actually} stays below $13\%$ and
$9\%$ at every position. A reflection marker is therefore associated with a weaker trial yield, while the thinking-close signal demonstrates a higher observed yield. This changing signal mixture partly explains why the pooled marker lane of Figure~\ref{fig:two-triggers}
decays: after $400$ tokens its events are increasingly textual markers rather
than close signals. The bottom panel of Figure~\ref{fig:signal-comparison} shows
the second half of the window argument. Up to $700$ tokens the marker lane loses
yield because its confidence test passes less often, from $29\%$ at $100$--$200$ tokens down to
$13\%$ at $600$--$700$, not because the answers get worse. Beyond $800$ tokens
both fail at once: the confidence test keeps passing and the mean trial confidence stays flat
near $0.80$, but the share of passed tests whose answer is correct collapses from
$78\%$ to $45\%$. This lower late pooled accepted-answer quality despite stable mean confidence motivates retiring trials.

%% file: case_studies_appendix.tex
\section{Case studies}
\label{app:case-studies}

\providecommand{\globestop}[1]{\textcolor{worse}{\textbf{[#1]}}}
\providecommand{\caseok}{\textcolor{better}{\textbf{(correct)}}}
\providecommand{\casewrong}{\textcolor{worse}{\textbf{(wrong)}}}

\newmdenv[
  linewidth=0.4pt,
  linecolor=black,
  skipabove=4pt,
  skipbelow=4pt,
  innertopmargin=2.5pt,
  innerbottommargin=2.5pt,
  innerleftmargin=4pt,
  innerrightmargin=4pt
]{globecase}

This appendix walks through five items.
The four Qwen items come from the GLOBE runs behind
Table~\ref{tab:all-results}, and the Gemma item from a GLOBE run on the same
split. Reasoning excerpts are
verbatim and trimmed with [\dots]; the red marker shows where GLOBE ends the
main stream and what it observes there. $J$ is the number of completed segments,
$K$ the reversal count, and $Q$ the net-displacement ratio at the stop.

\subsection{Trial-answer exit at a thinking-close signal}
\label{app:case-trial}

\begin{globecase}\footnotesize\raggedright
\textbf{BFCL v4 Live, Qwen} (\texttt{live\_multiple\_829-178-4}).\quad
\textbf{User:} ``Can you play `Bohemian Rhapsody' in the kitchen for me?''\\[3pt]
\textbf{GLOBE reasoning} (109 retained tokens):
\emph{The user wants to play the song `Bohemian Rhapsody' in the kitchen. I need
to use the \texttt{Music\_3\_PlayMedia} function. Parameters required:
track: ``Bohemian Rhapsody''; device: ``Kitchen''; artist: optional, default
``any''; album: optional, default ``any''. I will call
\texttt{Music\_3\_PlayMedia} with \texttt{track="Bohemian Rhapsody"} and
\texttt{device="Kitchen"}. All}\ \globestop{GLOBE stops: \texttt{</think>}
enters the top-5 at $J=3$; trial
\texttt{\{Music\_3\_PlayMedia(track="Bohemian Rhapsody", device="Kitchen")\}},
$q=0.945>\tau_{\mathrm{conf}}$}\\[3pt]
\textbf{GLOBE answer:} \texttt{Music\_3\_PlayMedia(track="Bohemian Rhapsody",
device="Kitchen")} \caseok\\[3pt]
\textbf{Full CoT:} writes the same call in its first paragraph, then keeps
confirming it: \emph{``[Final Output Generation] $\rightarrow$
\{"name": "Music\_3\_PlayMedia", \dots\} Done. Proceeds. (Self-Correction/Verification
during thought) Wait, I should check if there are any other constraints. None.
Proceeds. [\dots]''} The loop runs until the token budget is exhausted, and no
call is emitted \casewrong
\end{globecase}

\smallskip
The model's own thinking-close token enters the top-5 candidates after three
segments, when the call and both arguments are already fixed; the trial branch
confirms it with confidence $0.945$ and the continuation emits exactly that
call after 109 reasoning tokens. Full CoT has the same call in hand at the same
point but never commits to it: it restates it until the budget runs out.

\subsection{Trajectory stop on an argument that keeps flipping}
\label{app:case-live-trajectory}

\begin{globecase}\footnotesize\raggedright
\textbf{BFCL v4 Live, Qwen} (\texttt{live\_multiple\_455-145-6}).\quad
\textbf{User:} ``Search for free entry gardens suitable for a short trip in
London, UK?''\\[3pt]
\textbf{GLOBE reasoning} (311 retained tokens; one trial, \texttt{\{London,
UK\}} at $q=0.769$, rejected):
\emph{[\dots] \texttt{category}: ``Park'' or ``Nature Preserve'' might be closest to
``gardens'', but the user said ``gardens''. [\dots] I'll use ``Park'' or
``dontcare'' and filter by free\_entry. Actually, ``Park'' is the best match for
gardens. Or I can just leave category as ``dontcare'' and rely on free\_entry.
Let's stick to ``Park'' or ``dontcare''. I'll use ``dontcare'' to be safe, or maybe
``Park''. Actually, gardens are often categorized as Parks. I'll use ``Park''.
Wait, the prompt says ``gardens''. I'll just use ``dontcare'' for category to be
safe, or ``Park''. Let's use ``Park''. [\dots] Let's check the parameters:}\
\globestop{GLOBE stops: $J=10$, $K=8$, $Q=0.127\le\tau_Q$}\\[3pt]
\textbf{GLOBE answer:} \texttt{Travel\_1\_FindAttractions(location="London, UK",
category="Park", free\_entry="True", good\_for\_kids="dontcare")} \caseok\\[3pt]
\textbf{Full CoT:} settles on the same arguments early, then alternates between
\emph{``[Output Generation] $\rightarrow$ Tool Call.''} and \emph{``[Final Output
Generation] $\rightarrow$ Tool Call. [\dots] Done. Proceeds.''} until the token
budget is exhausted, and no call is emitted \casewrong
\end{globecase}

\smallskip
The trace has already collected every argument; what remains is a choice between
two enum values that it revisits sentence after sentence. On the unit sphere
these sentences alternate between two nearby directions, so every new segment
reverses the previous step while the path makes almost no net progress: after
ten segments the reversal count reaches $K_{\mathrm{osc}}=8$ with a
net-displacement ratio of $0.127$, and GLOBE ends the stream after 311 reasoning
tokens. The continuation commits to \texttt{"Park"} and returns a correct call.
Full CoT resolves the same dilemma, but then loops on its own output and never
closes; this is the failure that the trajectory stop is designed to cut.

\subsection{Phase switch and a trajectory stop that rescues an answer}
\label{app:case-gsm8k}

\begin{globecase}\footnotesize\raggedright
\textbf{GSM8K, Qwen} (test item 100).\quad \textbf{Question:} ``Jerome had 4
friends who came to visit him on a certain day. The first friend pressed on the
doorbell 20 times before Jerome opened, the second friend pressed on the
doorbell 1/4 times more than Jerome's first friend. The third friend pressed on
the doorbell 10 times more than the fourth friend. If the fourth friend pressed
on the doorbell 60 times, how many doorbell rings did the doorbell make?''
(gold: 175)\\[3pt]
\textbf{Trial phase:} three thinking-close trials at $J=2,4,5$ return
\texttt{\{10\}}, \texttt{\{135\}}, and \texttt{\{135\}} with confidence $0.587$,
$0.709$, and $0.755$; all are rejected. The reversal count then passes
$K_{\mathrm{phase}}=4$ and trials are retired.\\[3pt]
\textbf{GLOBE reasoning} (659 retained tokens): \emph{[\dots] Sum = 20 + 25 +
70 + 60 = 175. 4. Check for Ambiguities/Traps: [\dots] ``10 times more than''
usually means original + 10. Wait, ``10 times more than'' can be ambiguous. It
could mean 10 * F4 = 600, or F4 + 10 = 70. [\dots] Actually, ``10 times more than''
strictly means multiplication by 10 (or 11x depending on interpretation), but in
elementary math problems, it's often a typo for ``10 more than''. Let's check
typical patterns in these datasets. Often, ``X times more than'' means addition if
X is small, but here it says ``10 times more than''.}\ \globestop{GLOBE stops:
$J=15$, $K=13$, $Q=0.096$}\\[3pt]
\textbf{GLOBE answer:} \emph{``The third friend pressed it 10 times more than the
fourth friend: 60 + 10 = 70 times. Total doorbell rings = 20 + 25 + 70 + 60 =
175.''} \caseok\\[3pt]
\textbf{Full CoT answer:} \emph{``The third friend pressed 10 times more than the
fourth, which is 10 * 60 = 600 times. The total number of doorbell rings is 20 +
25 + 600 + 60 = 705.''} \casewrong
\end{globecase}

\smallskip
This item exercises all three components. The early trials are launched while
the computation is unfinished and return wrong answers ($10$ and $135$); their
confidence stays below $\tau_{\mathrm{conf}}$, so none of them ends the stream. Once the
reversal count reaches $K_{\mathrm{phase}}$, trials are retired, and the
trajectory records the trace circling between two readings of ``10 times
more than''. Thirteen reversals with a net-displacement ratio of $0.096$ end the
stream while the trace is still at its first, correct computation, and the
continuation answers $175$. Full CoT keeps deliberating, settles on the other
reading, and answers $705$: an instance of the rescue measured in
Figure~\ref{fig:stop}a.

\subsection{Trajectory stop on a multiple-choice trace}
\label{app:case-csqa}

\begin{globecase}\footnotesize\raggedright
\textbf{CSQA, Qwen} (\texttt{validation-000124}).\quad \textbf{Question:}
``From where do aliens arrive?'' (A) outer space (B) weekly world news
(C) roswell (D) universe (E) mars (gold: A)\\[3pt]
\textbf{GLOBE reasoning} (345 retained tokens, no trial): \emph{[\dots] (A) outer
space: This is the general location where extraterrestrial life is hypothesized
to exist. [\dots] (D) universe: While true in a broad sense, ``outer space'' is the
specific medium through which they travel to Earth. [\dots] While they might come
from Mars (E) or the Universe (D), ``outer space'' (A) is the most encompassing
and standard answer [\dots] However, let's look at the structure of the question.
It looks like a common sense or trivia question.}\ \globestop{GLOBE stops:
$J=10$, $K=8$, $Q=0.135\le\tau_Q$}\\[3pt]
\textbf{GLOBE answer:} \texttt{\textbackslash boxed\{A\}} \caseok
\end{globecase}

\smallskip
The trace reweighs options already ruled out, causing the reversal count to grow while net displacement stays small ($K=8$, $Q=0.135$ after 10 segments). GLOBE stops after 345 reasoning tokens with the correct option.


\subsection{An early trial on Gemma}
\label{app:case-gemma}

\begin{globecase}\footnotesize\raggedright
\textbf{GSM8K, Gemma} (test item 298).\quad \textbf{Question:} ``Carmen goes to
an auction to win an antique desk. She accepts the opening bid of \$200 and
continues bidding until she wins. The bids on the desk rise by \$50 each time and
3 other people each bid once. Carmen bids after each of the 3 other people and
eventually wins. How much money, in dollars, does the desk cost her?'' (gold:
500)\\[3pt]
\textbf{GLOBE reasoning} (105 retained tokens): \emph{[\dots] Opening bid: \$200
(Carmen accepts this). Bids rise by \$50 each time. There are 3 other people who
each bid once. Carmen bids *after* each of the 3 other people. Carmen eventually
wins. Goal: Find the final cost of the desk. Initial state: Carmen accepts the
opening bid of \$200.}\ \globestop{GLOBE stops: \texttt{Wait} at $J=2$; trial
\texttt{\{350\}}, $q=0.925>\tau_{\mathrm{conf}}$}\\[3pt]
\textbf{GLOBE answer} (continuation from the retained prefix): \emph{``[\dots] The
second other person bids: \$300 + \$50 = \$350. Carmen bids after the second
person: \$350 + \$50 = \$400. The third other person bids: \$400 + \$50 = \$450.
Carmen bids after the third person to win: \$450 + \$50 = \$500. The desk costs
her \$500.''} \caseok\\[3pt]
\textbf{Full CoT:} exhausts its 8{,}192-token budget without a final answer
\casewrong
\end{globecase}

\smallskip
The trial answers readiness, not the value of the final answer: its numeral
\texttt{\{350\}} is only a probe and is not inherited by the continuation, which
produces the visible answer afterwards from the retained prefix under the task's
own format. Here the prefix has already laid out every rule of the auction, and
the continuation carries the computation through to \$500 in the visible answer,
while Full CoT on the same item never reaches an answer inside its budget.